\documentclass[11pt]{article}

\usepackage{acl}

\usepackage{times}
\usepackage{latexsym}
\usepackage[T1]{fontenc}
\usepackage[utf8]{inputenc}
\usepackage{microtype}
\usepackage{inconsolata}
\usepackage{graphicx}
\usepackage{amsmath}
\usepackage{amssymb}

\usepackage{multirow}
\usepackage{makecell}
\usepackage{booktabs}

\usepackage{array}
\usepackage{siunitx}
\usepackage[table]{xcolor}

\title{Correct When Paired, Wrong When Split: Decoupling and Editing Modality-Specific Neurons in MLLMs}

\usepackage{tcolorbox}
\newtcolorbox{prompt}[1]{colback=gray!5,colframe=gray!35!black,fonttitle=\bfseries, title={#1}}

\newcommand{\ours}[1]{\textsc{DECODE}}
\newcommand{\eqcontrib}{%
\textsuperscript{\textcolor{orange}{\(\blacklozenge\)}}%
}
\newcommand{\blfootnote}[1]{%
  \begingroup
  \renewcommand{\thefootnote}{}%
  \footnotetext{#1}%
  \addtocounter{footnote}{-1}%
  \endgroup
}

\author{
 \textbf{Tingchao Fu\textsuperscript{1,}\eqcontrib},
 \textbf{Wenkai Wang\textsuperscript{2,}\eqcontrib},
 \textbf{Fanxiao Li\textsuperscript{1}},
 \textbf{Huadong Zhang\textsuperscript{1}},
\\
 \textbf{Jinhong Zhang\textsuperscript{1}},
 \textbf{Dayang Li\textsuperscript{1}},
 \textbf{Yunyun Dong\textsuperscript{1}},
 \textbf{Renyang Liu\textsuperscript{3}},
 \textbf{Wei Zhou\textsuperscript{4,}*}
\\
\\
 \textsuperscript{1}School of Information Science and Engineering, Yunnan University
 \\
 \textsuperscript{2}School of Software, Yunnan University, \textsuperscript{3}National University of Singapore
 \\
 \textsuperscript{4}School of Engineering, Yunnan University
\\
 \small{
   {\texttt{futingchao@stu.ynu.edu.cn, ryliu@nus.edu.sg, zwei@ynu.edu.cn}}
 }
}

\begin{document}

\maketitle

\blfootnote{\eqcontrib: Equal contributions. *: Corresponding author.}
\blfootnote{Data and code are available at \url{https://github.com/TingchaoFu/DECODE}.}

\begin{abstract}
Although Knowledge Editing provides an efficient mechanism for updating the knowledge of Multimodal Large Language Models (MLLMs), we find that current paradigms still suffer from an important yet remain underexplored issue: \textbf{\emph{editing decoupling failure}}, where entity-related knowledge can be updated when the model is triggered by multimodal inputs (text--image query pairs), however, it often reverts to outdated pre-edit facts when the paired inputs are split into unimodal ones. Our in-depth empirical analysis reveals that the entity knowledge in MLLMs is not stored as a unified representation, but is instead distributed across disentangled \textbf{\emph{modality-specific}} pathways. As a result, updates biased toward multimodal queries fail to propagate effectively to unimodal circuits. To bridge this gap, we propose DECODE, which explicitly disentangles and localizes modality-specific neuron groups for targeted knowledge. Extensive experiments demonstrate that DECODE consistently achieves effective knowledge updates under different modality triggers, thereby mitigating \textit{editing decoupling failures}.
\end{abstract}

\section{Introduction}
\label{sec:intro}
Multimodal Large Language Models (MLLMs)~\citep{instructblip, blip2, gpt4, qwen3} have demonstrated remarkable capabilities such as cross-modal understanding \cite{understand,liu2025}, reasoning \cite{reasoning, Haoran}, and knowledge-intensive tasks \citep{drift,see}. However, the static parametric knowledge substantially limits the development in real-world scenario that inquire up-to-date information \citep{ke, cmie, li2025imrrf}. Given the high costs of retraining or fine-tuning, Knowledge Editing is considered an efficient alternative.~\citep{mend,Hallucinations,ccs}, which aims to precisely modify a small subset of parameters to update intrinsic knowledge while largely preserving unrelated knowledge.
\begin{figure}
  \centering
  \includegraphics[width=0.99\linewidth]{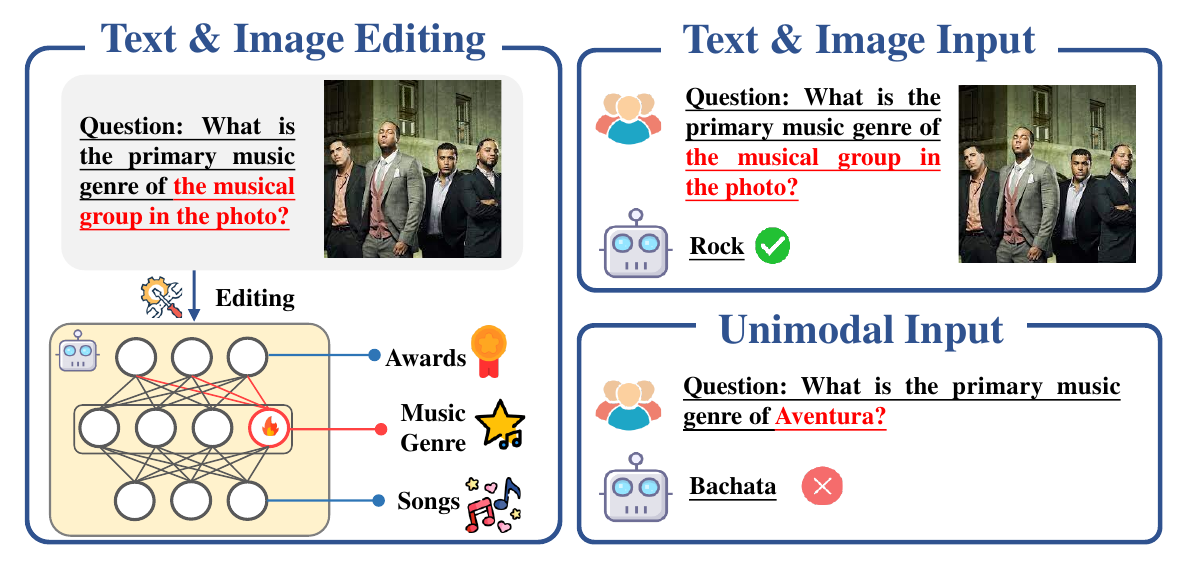}
  \caption{Illustration of \textit{\textbf{editing decoupling failure.}} While prior knowledge editing methods successfully update facts when presented with multimodal inputs, they suffer from editing decoupling failure: once the input is decoupled into a unimodal format, the MLLM reverts to providing erroneous or outdated information.}
  \label{fig:motivation}
 
\end{figure}

Most existing knowledge editing methods for MLLMs inherit LLM-centric editing paradigms and are typically optimized for a specific \textit{text-conditioned} query \cite{rome,li2024pmet,grace}. These approaches treat whether the final answer is corrected as the primary editing objective and evaluation criterion \cite{mmke}, but \textit{lacking dedicated constraints to enforce the alignment between the multimodal input content and the target entity being edited}. Therefore, a few literature recently have started to analyze MLLMs specific features and attempt to parse and edit activation patterns induced by multimodal inputs \cite{unike, aaaiedit, ike}. However, they often implicitly assume that cross-modal knowledge is shared and transferable: once an update is successfully applied under multimodal inputs, it should also hold under unimodal inputs \cite{finding}.

Nevertheless, our empirical study reveals a key challenge faced by above paradigm, which we term \textit{editing decoupling failure}. Even the post-edited MLLMs appears output correctly on the text-image paired queries, the model can still revert to pre-edit knowledge by using the text-only input (as shown in Figure~\ref{fig:motivation}). This suggests that in MLLMs, the same factual attribute can be elicited by different input compositions, and the entity can be referred to and grounded in different ways (explicit entity names vs. visual references). Consequently, multimodal knowledge updates may not automatically and consistently propagate across modality-specific pathways and their fusion mechanisms. Correcting the output under one input form does not guarantee consistent updates across modalities or across different forms of entity reference.

Specifically, we conduct an in-depth analysis of neuron activation patterns within MLLMs and find that the knowledge representation of the same entity exhibits significant modality specificity. Two relatively decoupled groups of neurons are observed: (1) One primarily responds to multimodal features. In this case, text--image pairs are inputted to MLLMs, and the entity can exist in both all or just in the image; (2) The other is more dedicated to unimodal features, and the entity appears only in the text input. This disentangled storage mechanism explains a key limitation of existing editing methods: they are typically activated and optimized under multimodal input settings, and thus tend to update multimodal-responsive neurons while overlooking another group of neurons responsible for unimodal reasoning and knowledge retrieval. 

Motivated by the above analysis, we propose 
\textit{\textbf{\ours{}}: \textbf{D}isentangling \textbf{E}ntity knowledge via \textbf{C}ross-m\textbf{O}dal critical-neuron \textbf{D}emarcation for \textbf{E}diting}, a novel framework designed to ensure knowledge consistency across different modality pathways. DECODE explicitly decouples modality-specific neurons and performs targeted editing on distinct activation groups, thereby improving the consistency of cross-modal knowledge updates. Concretely, for a given entity: (1) DECODE constructs modality-specific queries from the original input and computes each neuron’s contribution score under the corresponding modality. Based on these scores, it groups and localizes modality-specific neurons. (2) We further introduce a decoupled two-stream editing strategy: it optimizes the offsets of visual-aware neurons, and updates the offsets of textual-aware neurons via text-only queries, respectively.  In this way, DECODE effectively mitigates and resolves the \emph{editing decoupling failure}, enabling updated knowledge consistently propagate across modalities.

Extensive evaluations across multiple MLLM architectures demonstrate that DECODE significantly outperforms existing baselines, achieving superior reliability, generality and locality in both multimodal and decoupled unimodal settings.

\section{Preliminary}
\label{sec:pre}
\paragraph{Multimodal Large Language Model.} An MLLM $\mathcal{M}(\cdot)$ typically consists of three components \cite{mllm}: an \textit{LLM backbone}, which serves as the core reasoning module \cite{llama,vicuna}; a \textit{visual encoder}, which extracts representations from the input image \cite{clip}; and a \textit{projector or Q-Former}, which maps visual representations into the language embedding space to facilitate multimodal alignment \cite{minigpt}. A response can be generated with the following process:
\begin{equation}
Response = \mathcal{M}(\mathcal{I}, \mathcal{P}),
\end{equation}
where the visual encoder first extracts high-level features from the input image $\mathcal{I}$, and the projector further maps these visual features into the token embedding space of the LLM. The resulting visual tokens are then concatenated with the text tokens of the prompt $\mathcal{P}$, enabling the LLM backbone to perform joint multimodal reasoning and generate the final response.

\paragraph{Knowledge Editing.} Knowledge editing aims to update specific factual associations stored in a model without retraining it from scratch \cite{li2024pmet,emmet}. Following common formulations, factual knowledge can be represented as triples $(s, r, o)$, where $s$ denotes the subject, $r$ the relation, and $o$ the object. 
Given an edit target that replaces $o$ with $o^{*}$ under the same $(s, r)$, the objective is to obtain an edited model $\tilde{\mathcal{M}}$ such that:
\begin{equation}
    o^{*} = \tilde{\mathcal{M}}(s, r).
\end{equation}
In the multimodal setting, the subject $s$ is grounded in both an image $\mathcal{I}$ and a textual instruction $\mathcal{P}$.

\begin{figure*}[!t]
    \centering
    \includegraphics[width=0.99\linewidth]{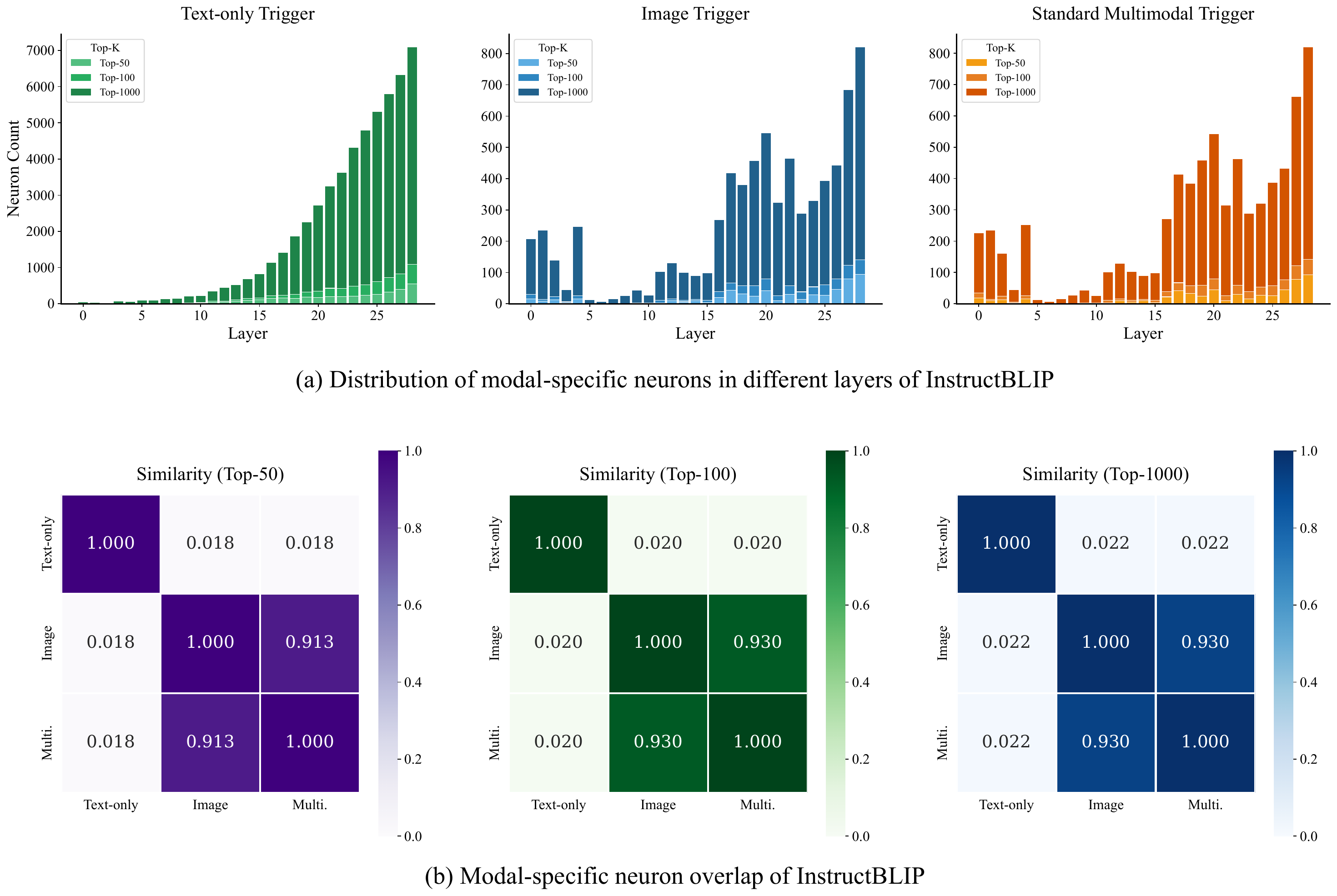}
    \caption{\textbf{Layer-wise distribution and modality overlap of critical neurons in InstructBLIP.} The analysis considers high-contribution neurons at different granularities, including the top-50, top-100, and top-1000 neurons ranked by contribution scores.}
    \label{fig:neuron}
  
\end{figure*}

To achieve the above goals, we modify neurons in the feed-forward network (FFN) layers of the \emph{LLM backbone}, which have been shown to store substantial amounts of factual knowledge~\cite{rome,memit,alphaedit}.
Specifically, for an input token sequence $\mathcal{X}$ (including projected visual tokens and text tokens), the hidden state at layer $l$ is updated as:
\begin{equation}
    \mathbf{h}^{l} = \mathbf{h}^{l-1} + \mathbf{a}^{l} + \mathbf{m}^{l},
\end{equation}
\begin{equation}
\label{eq:ffn}
\mathbf{m}^{l} = \mathbf{W}_{out}^{l}\,\sigma\!\left(\mathbf{W}_{in}^{l}\,\gamma\!\left(\mathbf{h}^{l-1}+\mathbf{a}^{l}\right)\right),
\end{equation}
where $\mathbf{a}^{l}$ and $\mathbf{m}^{l}$ denote the outputs of the attention block and the FFN block, respectively; $\mathbf{W}_{in}^{l}$ and $\mathbf{W}_{out}^{l}$ are the two FFN projection matrices; $\sigma(\cdot)$ is the activation function; and $\gamma(\cdot)$ denotes layer normalization. 
For simplicity, we define $\mathbf{O}^l=\sigma\!\left(\mathbf{W}_{in}^{l}\gamma\!\left(\mathbf{h}^{l-1}+\mathbf{a}^{l}\right)\right).$

\textbf{Neuron Contribution Score. }
Prior work~\cite{finding} shows that neurons with higher contribution scores are more likely to be activated during generation. Therefore, we first identify these high-contribution neurons as targets for subsequent editing. Formally, for a given concept token $t$, the contribution score $s_{i,t}^{l}$ of the $i$-th neuron $u_{i}$ at layer $l$ is defined as:
\begin{equation}
s_{i,t}^{l} = \mathbf{Q}^{l}(i,t),
\end{equation}
where $\mathbf{Q}^l = (\mathbf{W}_u \mathbf{W}_{\text{out}}^l) \odot \mathcal{T}(\mathbf{O}_{-1}^l) \in \mathbb{R}^{d_m \times v}$ represents the logit-space projection of the neuron activations, $d_{m}$ denotes the FFN intermediate size, and $v$ denotes the vocabulary size. Here, $\mathbf{W}_u$ denotes the unembedding matrix, $\mathbf{W}_{\text{out}}^l$ is the output projection matrix of the $l$-th layer, $\mathbf{O}_{-1}^l$ represents the activation at the final token position, and $\odot$ denotes the element-wise (Hadamard) product. A higher magnitude of $s_{i,t}^{l}$ signifies that the neuron $u_i$ exerts a more significant influence on the prediction of the target concept.

\section{Exploratory Study}
Building upon the existing multimodal knowledge editing benchmark MMKE~\cite{mmke}, we further investigate whether a model that successfully edits knowledge about the same entity under \textit{multimodal compositional triggers} can still maintain consistent and stable knowledge-updating behavior after modality decoupling.

\subsection{Dataset Construction}
\label{sec:data}
In the original MMKE, the target knowledge is typically triggered by \textit{joint cues from texts and images}. To study knowledge editing consistency under modality splitting, we adapt each sample into three evaluation settings that selectively activate different modality pathways (details are in Appendix~\ref{sec:data_details}):
\paragraph{Text-only trigger}: We query the edited knowledge using only text, without any visual input, to assess whether the model can retrieve the updated fact through a purely linguistic pathway.
\paragraph{Image trigger}: We provide both the image and text, but the text contains no explicit entity name; instead, it refers to the target entity via indicative QA-style expressions (e.g., “the object/person in the image”). This setting evaluates whether the model can rely on visual grounding to identify the entity and trigger the corresponding knowledge.
\paragraph{Standard multimodal trigger}: We provide both the image and text that explicitly contains entity information, serving as a standard setting with redundant cues, and use it to compare editing effectiveness and stability under compositional inputs.

\subsection{Modality-specific Neuron Decoupling and Locating}
\label{sec:decoupling}
Based on the constructed three knowledge-triggering variants, we compute the corresponding neurons' contribution scores under each evaluation setting following the procedure described in Section~\ref{sec:pre}, enabling the decoupling and localization of modality-specific neurons. We then rank all neurons in the \emph{LLM backbone} by their contribution scores and select the top-$k$ neurons with the highest scores ($k \in {50, 100, 1000}$), yielding the following three decoupled neuron sets:

\textbf{(1) Text neurons set ($U_t$)}: Obtained under the Text-only trigger setting, capturing the neurons that respond when the knowledge is triggered solely through the textual modality.

\textbf{(2) Visual neurons set ($U_v$)}: Obtained under the Image trigger setting, capturing the neurons that respond when the same knowledge is triggered by visual cues.

\textbf{(3) Multimodal neurons set ($U_m$)}: Obtained under the Multimodal trigger setting, identifying the neuron regions involved when entity knowledge is jointly activated by both textual and visual cues.

Based on this, we quantify neuron contribution scores under the three triggering settings and reveal the overlap among the identified sets of critical neurons, as shown in Figure~\ref{fig:neuron}.

\subsection{Analysis}
\label{sec:ana}
In Figure~\ref{fig:neuron}, we present an analysis of the activated regions of InstructBLIP~\cite{instructblip} under decoupled modalities (see Appendix~\ref{sec:llava} for more models). Building upon these results, our
observations are as follows:

\textbf{(1) The same entity knowledge exhibits modality-specific response regions under different triggers.} As shown by the similarity matrix in Figure~\ref{fig:neuron}b, the overlap between the text neuron set ($U_t$) and the visual neuron set ($U_v$) is extremely small: only 0.018 for Top-50 and 0.022 even for Top-1000. More notably, the multimodal neuron set ($U_m$) is highly similar to the visual set ($U_v$) (about 0.930), yet is almost disjoint from the text set ($U_t$). This suggests that once visual stimuli are introduced, the model tends to retrieve and activate knowledge along a vision-dominated pathway, thereby largely re-routing away from the key neurons relied upon by purely language-triggered queries.

\textbf{(2) Different modalities can induce hierarchical re-routing of knowledge, modulated by model architecture.} For the same entity knowledge, different triggering settings lead to markedly different layer-wise distributions of critical neurons. Under the text-only trigger, the high-contribution neuron set of InstructBLIP ($U_t$) is concentrated in deeper layers, consistent with prior findings on knowledge localization~\cite{rome, memit}. In contrast, under the image trigger and multimodal trigger, the activated regions shift noticeably toward shallower layers, suggesting that vision-related cues may intervene at earlier backbone layers. Moreover, this hierarchical re-routing exhibits architecture-dependent patterns: for example, LLaVA and Qwen-VL still show predominantly deep-layer activations under three triggers, indicating that the cross-modal interface and fusion timing can modulate where the same knowledge is activated across modalities (see Appendix~\ref{sec:llava}).

\begin{figure*}
    \centering
    \includegraphics[width=0.99\linewidth]{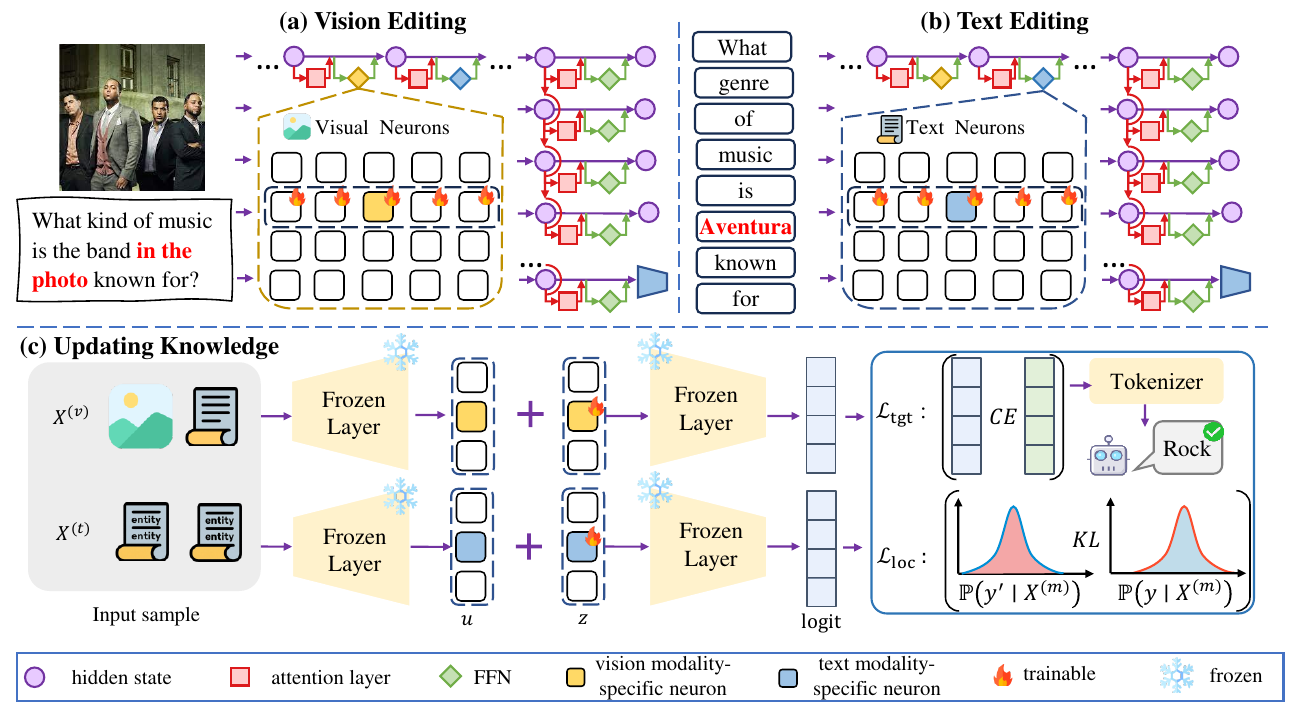}
    \caption{\textbf{Overview of \ours{}.}  (a-b) The framework explicitly targets decoupled visual ($U_v$) and textual ($U_t$) neurons. (c) A two-stream process injects learnable offsets $z$ into these FFN neurons, optimizing target ($\mathcal{L}_{tgt}$) and locality ($\mathcal{L}_{loc}$) losses to synchronize cross-modal knowledge.}
    \label{fig:DCODE}
\end{figure*}

\begin{tcolorbox}
[colback=black!2!white,colframe=white!50!black,boxrule=0.5mm]
\textit{\textbf{Key Finding:} Knowledge in MLLMs is sequestered into modality-specific pathways rather than being integrated into a unified, entity-centric representation.}
\end{tcolorbox}

This key finding leads to the phenomenon of \emph{editing decoupling failure}. Under multimodal knowledge triggering, parameter updates primarily act on the multimodal critical neuron set $U_m$ (which largely overlaps with $U_v$). However, since $U_t$ remains significantly decoupled from $U_m/U_v$ both spatially and functionally across different choices of $k$, the updated knowledge is more likely to be written into—and thus “trapped” within—the vision-dominated pathway. Consequently, when the model later re-triggers the same knowledge in a text-only manner, the retrieval process is still routed to the stale, unedited parameters in $U_t$, resulting in decoupled editing failure.

\section{Methodology}
\label{sec:decode}
To address \emph{editing decoupling failure} in MLLMs, we propose \textit{\textbf{\ours{}}}: \textbf{D}isentangling \textbf{E}ntity knowledge via \textbf{C}ross-m\textbf{O}dal critical-neuron \textbf{D}emarcation for \textbf{E}diting. The overall framework is shown in Figure~\ref{fig:DCODE}, \ours{} first identifies entity-related, modality-specific critical neuron sets for the text and image modalities, and then performs two-stream editing over these distinct response sets to achieve consistent knowledge updates across modalities.

\paragraph{Modality-specific neuron decoupling.}
We first obtain modality-specific neuron sets $U_t$ and $U_v$, following the same construction as in our empirical study (Section~\ref{sec:decoupling}). Specifically, $U_t$ and $U_v$ consist of the top-$k$ neurons with the highest contribution to activating the target knowledge under the text-trigger input $X^{(t)}$ and the image-trigger input $X^{(v)}$.

\paragraph{Neuron-wise parameter offsets.}
For each neuron $u \in U_t \cup U_v$, we introduce a learnable offset vector $z_u$ and inject it into the FFN parameter row corresponding to neuron $u$ via element-wise addition. Following the FFN definition in Eq.~(\ref{eq:ffn}), this offset is applied to the output projection $W^{l}_{\text{out}}$:
\begin{equation}
\label{eq:DECODE_inject}
\tilde{W}^{l}_{\text{out}}[u,:] \;=\; W^{l}_{\text{out}}[u,:] \;+\; z^{l}_{u},
\forall u \in U_t \cup U_v,
\end{equation}
where $z^{l}_{u}$ has the same dimension as one row of $W^{l}_{\text{out}}$.
The edited model parameters are denoted by $\theta'$ (i.e., $\theta$ with $\{z_u^l\}$ applied).

To ensure knowledge is effectively synchronized across decoupled pathways, DECODE performs editing through a two-stream paradigm. This approach targets the modality-specific neuron sets $U_v$ and $U_t$ separately, ensuring that updates to one modal stream do not interfere with the parameters of the other. While the editing involves two distinct stages, the execution order is flexible and can be adapted to different MLLM architectures (e.g., visual-first or textual-first).

\noindent\textbf{Modality-specific Updates.}
For each modality stream $m \in \{v, t\}$, DECODE optimizes the learnable offset vectors $\{z_u^l\}$ associated with the corresponding neuron set $U_m$. During the optimization of stream $m$, we only update the offsets within $U_m$ while keeping other parameters fixed.
This ensures that each stage of the two-stream process is surgically precise, updating only the functional rows of $W^{l}_{\text{out}}$ responsible for the targeted modality.

\noindent\textbf{Optimization Objective.} Each stream is optimized using a modality-specific objective that balances editing efficacy with knowledge locality. For a given stream $m$, the objective is:
\begin{equation}
\label{eq:edit_obj_stage}
\min_{\theta'}\mathcal{L}^{(m)} = \mathcal{L}_{\text{tgt}}\!\left(\theta'\right) + \lambda\mathcal{L}_{\text{loc}}\!\left(\theta'\right)
\end{equation}
where $X^{(m)}$ represents the modality-specific input. The target Loss $\mathcal{L}_{\text{tgt}}$ ensures the model correctly associates the new fact $o^{*}$ with the query $X^{(m)}$:
\begin{equation}
\label{eq:DECODE_tgt}
\mathcal{L}_{\text{tgt}}\!\left(\theta'\right)
=
-\log \mathbb{P}_{\theta'}    \!\left(o^{*}\mid X^{(m)} \right),
\end{equation}
To mitigate collateral damage to unrelated knowledge (a critical requirement for preserving the model's general capabilities), we employ a locality loss $\mathcal{L}_{\text{loc}}$ based on KL-divergence:
\begin{equation}
\label{eq:DECODE_loc}
\mathcal{L}_{\text{loc}}\!\left(\theta'\right)
=
D_{\mathrm{KL}}
\big(
\mathbb{P}_{\theta'}(y'\mid X^{(m)}) \parallel
\mathbb{P}_{\theta}(y\mid X^{(m)})
\big),
\end{equation}
where $y$ and $y'$ denote the output logits of the original and edited MLLM under the modality input $X^{(m)}$, respectively.
This penalty ensures that the predictive distribution of the edited model remains consistent with the previous state for non-target knowledge within the same modality. 
By employing this two-stream refinement, DECODE forces the knowledge update to propagate through both the visual and textual pathways , directly addressing the editing decoupling failure observed in our exploratory study.

\begin{table*}[t]
\renewcommand\arraystretch{1.1}
\setlength{\tabcolsep}{5pt}
\small
\begin{center}
\scalebox{0.94}{
\begin{tabular}{c|c|cccc|cccc|cc|cc}
\hline \hline
\multirow{3}{*}{\textbf{Model}} &
\multirow{3}{*}{\textbf{Method}} &
\multicolumn{4}{c|}{\textbf{Reliability}} &
\multicolumn{4}{c|}{\textbf{Generality}} &
\multicolumn{2}{c|}{\textbf{Locality}} &
\multicolumn{2}{c}{\textbf{Avg.}} \\
\cline{3-14}
& & \multicolumn{2}{c|}{\textbf{T}} & \multicolumn{2}{c|}{\textbf{M}} & \multicolumn{2}{c|}{\textbf{T}} & \multicolumn{2}{c|}{\textbf{M}} & \multirow{2}{*}{\textbf{T}} & \multirow{2}{*}{\textbf{M}} & \multirow{2}{*}{\textbf{EM}} & \multirow{2}{*}{\textbf{SM}} \\
\cline{3-10}
& & \textbf{EM} & \textbf{SM} & \textbf{EM} & \textbf{SM} & \textbf{EM} & \textbf{SM} & \textbf{EM} & \textbf{SM} & & & & \\
\hline

\multirow[c]{5}{*}{InstructBLIP}
& FT    & 98.24 & 97.78 & 99.39 & 99.88 & 96.23 & 96.05 & 99.26 & 99.57 & 32.05 & 20.95 & \underline{74.35} & \underline{74.38} \\
& SERAC & 28.10 & 27.39 & 99.63 & 99.14 & 15.93 & 15.24 & 99.63 & 99.14 & 100.00 & 99.61 & 73.82 & 73.42 \\
& IKE   & 10.83 & 66.07 &  3.83 & 63.23 &  6.37 & 56.76 &  4.30 & 64.03 &  2.48 &  1.63 &  4.91 & 42.37 \\
& FiNE  & 68.33 & 72.40 & 86.46 & 90.53 & 58.30 & 62.93 & 82.34 & 87.31 & 40.12 & 48.80 & 64.06 & 67.02 \\
& DECODE  & 96.27 & 96.79 & 91.69 & 92.16 & 83.58 & 84.81 & 91.08 & 91.42 & 46.11 & 45.45 & \textbf{75.70} & \textbf{76.12} \\
\hline

\multirow[c]{5}{*}{LLaVA}
& FT    & 99.72 & 99.75 & 99.98 & 100.00 & 99.38 & 99.26 & 99.98 & 100.00 & 21.07 & 20.38 & \underline{73.42} & 73.41 \\
& SERAC & 26.55 & 28.62 & 99.58 & 98.89 & 13.97 & 16.59 & 99.58 & 98.89 & 100.00 & 99.56 & 73.21 & \underline{73.76} \\
& IKE   & 36.47 & 80.57 & 13.11 & 73.78 &  9.16 & 76.13 & 13.48 & 73.53 &  9.16 & 38.12 & 19.92 & 58.55 \\
& FiNE  & 92.74 & 91.86 & 97.01 & 94.88 & 88.10 & 87.91 & 96.95 & 94.32 & 27.96 & 34.11 & 72.81 & 71.84 \\
& DECODE  & 94.49 & 93.27 & 91.05 & 91.36 & 86.55 & 87.47 & 88.54 & 90.43 & 54.31 & 55.77 & \textbf{78.45} & \textbf{78.77} \\
\hline

\multirow[c]{5}{*}{Qwen-VL}
& FT    & 98.21 & 98.95 & 98.83& 99.63 & 97.29 & 98.15 & 98.70 & 99.44 & 0.74 & 6.97 & 66.79 & 67.31\\
& SERAC    & 27.75 & 27.39 & 99.76& 99.07 & 16.07 & 16.22 & 99.76 & 99.07 & 96.18 & 97.92 & \underline{72.91} & \underline{72.64} \\
& IKE    & 84.68 & 93.21 & 25.16& 23.44 & 62.45 & 76.37 & 24.53 & 22.76 & 43.63 & 72.80 & 52.21 & 55.37\\
& FiNE    & 79.84 & 76.19 & 92.28& 88.83 & 66.93 & 64.03 & 90.35 & 86.92 & 18.79 & 18.15 & 61.06 & 58.82\\
& DECODE    & 92.69 & 91.98 & 97.59& 97.35 & 84.85 & 83.90 & 93.82 & 83.85 & 68.52 & 93.92 & \textbf{88.57} & \textbf{86.59} \\
\hline \hline
\end{tabular}}
\caption{Overall editing performance measured by \textbf{Exact Match (EM)} and \textbf{Semantic Match (SM)}. Locality is evaluated by calculating the cosine similarity between post-edit and pre-edit outputs to measure knowledge preservation. \textbf{Avg. EM} and \textbf{Avg. SM} represent the mean scores of EM and SM metrics integrated with Locality. Note that we employ free generation for all evaluations.}
\label{tab:main_results}
\end{center}
\end{table*}

\section{Experiments}
\subsection{Experimental Setting}
\paragraph{Models and Datasets.} We evaluate \ours{} on three MLLMs with distinct architectures: InstructBLIP-Vicuna-7B~\cite{instructblip}, LLaVA-v1.5-7B~\cite{llava} and Qwen-VL-7B~\cite{Qwen-VL}. They adopt different cross-modal alignment strategies: InstructBLIP introduces a Q-Former as a cross-modal bottleneck module, whereas LLaVA uses a Linear Projector with late-fusion alignment. We use the same data as in our empirical study; see Section~\ref{sec:data} for details.

\paragraph{Baselines. }
We select four representative categories of knowledge editing methods as baselines:
\textbf{(1) Parameter optimization:} \textit{Fine-tuning (FT)}, which updates model parameters via direct gradient descent on the editing objective. \textbf{(2) External-memory editing:} \textit{SERAC}~\cite{serac}, which stores edits in an external memory and uses a scope classifier to route inputs to an auxiliary model for editing.
\textbf{(3) In-context editing:} \textit{IKE}~\cite{ike}, a non-parametric method that performs knowledge updates via few-shot demonstrations in the prompt.
\textbf{(4) Neuron-level locating and editing:} \textit{FiNE}~\cite{fine}, which edits knowledge by localizing and updating specific functional neurons.

\paragraph{Evaluation Metrics.}
We adopt three evaluation metrics: \textit{Reliability}, \textit{Generality}, and \textit{Locality}. For each metric, we use two matching rules for scoring: Exact Match (EM) and Sub-string Match (SM).
\subsection{Main Results}
\label{exp: main_results}

We reported comparison results in Table~\ref{tab:main_results} and we concluded the following insights:
\textbf{\ours{} achieves superior cross-modal synchronization}. As shown in Table~\ref{tab:main_results}, baselines such as FiNE exhibit a severe editing decoupling failure: although they attain high reliability under the multimodal training setting ($M$-Reliability), their performance drops sharply when queried with pure text ($T$-Reliability). While the memory-based SERAC naturally achieves superior locality and reliability via its scope classifier, it demonstrates the most severe editing decoupling failure among baselines in multimodal contexts. This suggests that its classification mechanism, although effective for holistic inputs, fails to bridge the gap when modality triggers are split. \ours{}’s decoupled update strategy explicitly localizes and synchronizes these disjoint neural circuits, thereby resolving the decoupling bottleneck that hinders previous methods.

\textbf{Localization-based precision preserves model stability.}
Compared to optimization-based methods like FT, \ours{} demonstrates exceptional locality. While FT tends to overfit the edit fact and damage unrelated knowledge (dropping Locality to $20.38\%$ at LLaVA), DECODE maintains high surgical precision. By only modifying the functional neurons $U_t$ and $U_v$, \ours{} effectively updates factual knowledge while keeping the model’s broad generative capabilities and unrelated parametric memory intact.

\textbf{Architectural Robustness across Q-Former and Projector.} DECODE exhibits consistent performance gains across InstructBLIP, LLaVA and Qwen-VL. Despite their different cross-modal interfaces, our method adaptively localizes the respective modality-specific neurons. Notably, in InstructBLIP, where visual-related neurons are distributed in earlier layers, \ours{}'s fine-grained localization remains as effective as it is for LLaVA's deeper-layer concentration. This proves that targeting the functional neuron set is more robust than fixed-layer editing strategies.

\begin{table*}[t]
\renewcommand\arraystretch{1.1}
\setlength{\tabcolsep}{10pt}
\small
\begin{center}
\begin{tabular}{c|c|c|cc|cc|cc|c}
\hline \hline
\multirow{2}{*}{\textbf{Model}} &
\multirow{2}{*}{\textbf{Setting}} &
\multirow{2}{*}{\textbf{Stage}} &
\multicolumn{2}{c|}{\textbf{Reliability}} &
\multicolumn{2}{c|}{\textbf{Generality}} &
\multicolumn{2}{c|}{\textbf{Locality}} &
\multirow{2}{*}{\textbf{Avg.}} \\
\cline{4-9}
& & & \textbf{T} & \textbf{M} & \textbf{T} & \textbf{M} & \textbf{T} & \textbf{M} & \\
\hline

\multirow[c]{4}{*}{InstructBLIP}
& \multirow{2}{*}{Text First}  & Stage 1 & 88.71 & 60.70 & 65.08 & 60.54 & 57.54 & 59.06 & 65.27 \\
&                              & Stage 2 & 86.81 & 84.01 & 75.50 & 83.34 & 45.78 & 46.68 & \underline{70.35} \\
\cline{2-3}
& \multirow{2}{*}{Image First} & Stage 1 & 86.72 & 84.03 & 75.75 & 83.39 & 45.75 & 46.86 & 70.42 \\
&                              & Stage 2 & 96.27 & 91.69 & 83.58 & 91.08 & 45.45 & 46.11 & \textbf{75.70} \\
\hline

\multirow[c]{4}{*}{LLaVA}
& \multirow{2}{*}{Text First}  & Stage 1 & 65.94 & 25.83 & 53.72 & 25.50 & 72.50 & 72.41 & 52.65 \\
&                              & Stage 2 & 94.49 & 91.05 & 86.55 & 88.54 & 54.31 & 55.77 & \textbf{78.45} \\
\cline{2-3}
& \multirow{2}{*}{Image First} & Stage 1 & 63.88 & 88.52 & 58.51 & 83.66 & 46.73 & 51.37 & 65.45 \\
&                              & Stage 2 & 93.50 & 93.83 & 83.45 & 92.00 & 47.26 & 51.65 & \underline{76.95} \\

\hline \hline
\end{tabular}
\caption{\textbf{Editing performance comparison across different settings by Exact Match (EM).} We report text-only (T) and multimodal (M) scores for Reliability, Generality, and Locality, with the row-wise average (Avg.) in the final column. The results measured by Semantic Match (SM) are provided in Table~\ref{tab:sequence_SM} the Appendix.}
\label{tab:sequence}
\end{center}
\end{table*}

\subsection{Verification of Neuron Decoupling}
\label{exp: visualize neuron}

As illustrated in Appendix Figure~\ref{fig:vis}, the modality-specific neurons decoupled by DECODE effectively represent entity-related knowledge. Two models' neurons activated by textual queries show higher precision and concentration than those by visual inputs. This is because the main reasoning ability of MLLM is based on the internal knowledge of LLM, and MLLM is more sensitive to the textual entities. LLaVA maintains superior cross-modal consistency compare with InstructBLIP. Its projection-based design maps visual features directly into the LLM space, leading to a more uniform internal representation across modalities.

\subsection{Ablation Study}
\label{exp: ablation study}
To further evaluate the effectiveness of DECODE, this section investigates: \textbf{(i) the influence of editing locations}, comparing multimodal neurons $U_{m}$, union neurons (defined as the union set of $U_{t}$ and $U_{v}$), and the modality-specific localization in DECODE; and \textbf{(ii) the impact of editing order (image-first versus text-first)}.

\begin{figure}[!h]
    \centering
    \includegraphics[width=0.99\linewidth]{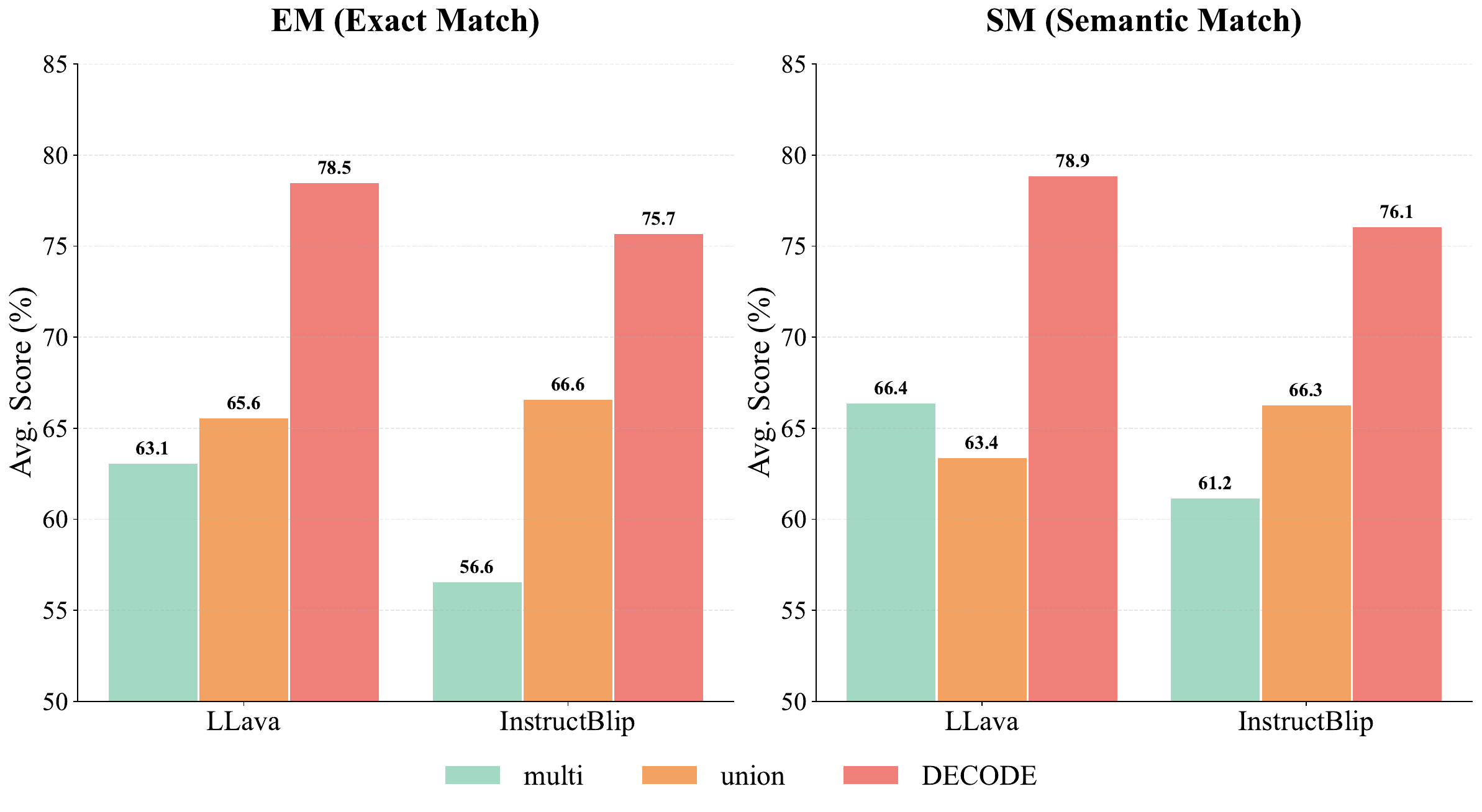}
    \caption{\textbf{Performance of editing at different location. } We report the mean scores of efficacy, locality, and generalization for LLaVA and InstructBLIP. DECODE consistently outperforms other localization strategies, including multimodal and union neurons.}
    \label{fig:ab1}
\end{figure}

\textbf{Impact of Editing Location.} As illustrated in Figure~\ref{fig:ab1}, the decoupled modality-specific neuron editing in DECODE consistently outperforms other localization strategies. Compared to editing only multimodal neurons, targeting both image and text neurons (i.e., the union set) yields superior average performance. This improvement can be attributed to the broader inclusion of relevant neurons, which bolsters both efficacy and generalization. However, this expansion comes with a notable trade-off: the increased number of modified neurons tends to degrade locality, which potentially leading to unintended interference with unrelated knowledge

\textbf{Impact of Editing Sequence.} We investigate the influence of the editing order—specifically, whether updating the visual-stream neurons before the textual-stream neurons or vice versa yields superior results. As reported in Table~\ref{tab:sequence}, we observe a distinct architectural sensitivity: LLaVA achieves higher performance with the Text-first strategy, whereas InstructBLIP exhibits a preference for the Image-first sequence.

This divergence is rooted in their distinct cross-modal alignment mechanisms. In InstructBLIP, the Q-Former's early and intensive integration of visual signals makes the visual stream a primary gateway for knowledge access; thus, an Image-first sequence is essential to establish a synchronized multimodal foundation. Conversely, LLaVA's linear projector maintains more disentangled modality pathways where the linguistic stream remains the dominant anchor for entity semantics. Consequently, a Text-first update better secures the core knowledge representation before aligning peripheral visual neurons.

\section{Related Work}
\paragraph{Knowledge Editing.} Early work formulates edits as updating specific factual associations and evaluates edits by \emph{reliability}, \emph{generality}, and \emph{locality} \cite{ke,mend}. A representative line of methods performs \emph{parameter-based} updates. ROME introduces a causal tracing strategy to localize where a fact is written and performs rank-one updates to precisely rewrite the corresponding association \cite{rome}. Beyond directly modifying model weights, \emph{external-memory} and \emph{non-parametric} editing paradigms avoid (or reduce) permanent parameter changes. SERAC stores edits in an external memory and routes inputs to an auxiliary module via a scope classifier, enabling reversible and modular updates \cite{serac}. In-context editing methods such as IKE instead “edit” knowledge at inference time through curated demonstrations, trading off permanence for flexibility \cite{ike}.

\paragraph{Neuron Analysis of MLLMs.}
Analyses of transformers suggest that factual knowledge can be localized to specific internal components and even specific neurons in feed-forward layers \cite{kv,rome,memit,emmet,mmedit,li2024pmet,alphaedit}. Such findings motivate \emph{neuron-level} editing, where updates are restricted to a small set of functionally relevant units. Recent studies start to explicitly characterize and manipulate multimodal activation patterns: FiNE \cite{fine} identifies high-contribution multimodal neurons and edits them to alter model outputs, while other approaches pursue collaborative or unified updates across modalities to multimodal editing \cite{finding}. Complementarily, VisEdit leverages attribution signals to correct visual knowledge and reduce erroneous associations \cite{aaaiedit}.

Despite these advances, most existing MLLM editing methods are optimized and validated primarily on \emph{paired} multimodal queries, implicitly assuming that once a fact is updated under a multimodal trigger, the update will transfer to unimodal queries that refer to the same entity. In contrast, our work reveals that the same entity knowledge can be sequestered into modality-specific pathways, causing \emph{editing decoupling failure} when the query is split into unimodal forms.

\section{Conclusion}
In this paper, we identify and characterize editing decoupling failure in MLLMs, a phenomenon where knowledge updates fail to propagate across different modality triggers. Our empirical analysis reveals that entity knowledge is sequestered into disentangled modality-specific pathways rather than a unified representation. To address this, we propose DECODE, which localizes and synchronizes these disjoint neural circuits through a two-stream editing strategy. Extensive experiments on multiple MLLM architectures demonstrate that DECODE significantly outperforms baselines in reliability and cross-modal consistency while preserving model locality. Our work highlights the necessity of considering modality-specific neuron distributions in multimodal knowledge editing.

\section*{Acknowledgments}
This work was supported by the Yunnan Research Project (Grant Nos. 202503AG380006, 202401AT070474, 202501AU070059, and 202403AP140021), the 17th Graduate Student Research and Innovation Program of Yunnan University (Grant No. KC-252512297), the National Natural Science Foundation of China (Grant Nos. 62562061, 62502422, and 62462067), and the Yunnan Provincial Department of Education Science Research Project (Grant Nos. 2025J0006, 2024J0010, and 2025J0007).

\section*{Limitations}
The known limitations and possible solutions of this work are as follows:
\begin{itemize}
\item Our method relies on internal activations (and typically gradients) to demarcate modality-specific critical neurons and inject neuron-wise offsets, which limits applicability to closed-source or API-only MLLMs. Future work could explore black-box approximations (e.g., probing-based surrogates) or retrieval and memory-based editing to support restricted-access models.
\item We mainly study entity-centric factual edits and diagnose \emph{editing decoupling} under modality-split prompts. This may not fully cover procedural knowledge, multi-hop reasoning, or long-context behaviors where knowledge is more distributed. Future work could extend DECODE to compositional edits and evaluate robustness under diverse prompting, longer dialogues, and sequential edits with stronger preservation constraints.
\end{itemize}

\section*{Ethical Considerations}
This work studies and mitigates \emph{editing decoupling failure} in MLLMs, aiming to make knowledge updates propagate consistently across different modality triggers, which may improve reliability when correcting outdated or erroneous facts and reduce persistent hallucinations tied to specific entities. 
We note, however, that knowledge editing is inherently dual-use: the same technique can be misused to inject targeted misinformation or stealthy behavioral changes. Although our goal is to improve factual consistency, responsible use requires access control, auditing, and reversibility (e.g., edit logs and post-edit monitoring) before deployment. Future work could incorporate stronger safety constraints (e.g., preservation-based guarantees), systematic red-teaming under modality-split prompts, and human-in-the-loop review for high-stakes edits.
\bibliography{custom}

\begin{thebibliography}{39}
\providecommand{\natexlab}[1]{#1}

\bibitem[{Achiam et~al.(2023)Achiam, Adler, Agarwal, Ahmad, Akkaya, Aleman, Almeida, Altenschmidt, Altman, Anadkat et~al.}]{gpt4}
Josh Achiam, Steven Adler, Sandhini Agarwal, Lama Ahmad, Ilge Akkaya, Florencia~Leoni Aleman, Diogo Almeida, Janko Altenschmidt, Sam Altman, Shyamal Anadkat, and 1 others. 2023.
\newblock Gpt-4 technical report.
\newblock \emph{arXiv preprint arXiv:2303.08774}.

\bibitem[{Bai et~al.(2023)Bai, Bai, Yang, Wang, Tan, Wang, Lin, Zhou, and Zhou}]{Qwen-VL}
Jinze Bai, Shuai Bai, Shusheng Yang, Shijie Wang, Sinan Tan, Peng Wang, Junyang Lin, Chang Zhou, and Jingren Zhou. 2023.
\newblock Qwen-vl: A versatile vision-language model for understanding, localization, text reading, and beyond.
\newblock \emph{arXiv preprint arXiv:2308.12966}.

\bibitem[{Caffagni et~al.(2024)Caffagni, Cocchi, Barsellotti, Moratelli, Sarto, Baraldi, Cornia, and Cucchiara}]{mllm}
Davide Caffagni, Federico Cocchi, Luca Barsellotti, Nicholas Moratelli, Sara Sarto, Lorenzo Baraldi, Marcella Cornia, and Rita Cucchiara. 2024.
\newblock The revolution of multimodal large language models: a survey.
\newblock \emph{arXiv preprint arXiv:2402.12451}.

\bibitem[{Chen et~al.(2025)Chen, Zhang, Wang, He, Wang, and Liu}]{aaaiedit}
Qizhou Chen, Taolin Zhang, Chengyu Wang, Xiaofeng He, Dakan Wang, and Tingting Liu. 2025.
\newblock Attribution analysis meets model editing: Advancing knowledge correction in vision language models with visedit.
\newblock In \emph{Proceedings of the AAAI Conference on Artificial Intelligence}, volume~39, pages 2168--2176.

\bibitem[{Cheng et~al.(2023)Cheng, Tian, Liu, Chen, Wang, Chen, and Zhang}]{mmedit}
Siyuan Cheng, Bozhong Tian, Qingbin Liu, Xi~Chen, Yongheng Wang, Huajun Chen, and Ningyu Zhang. 2023.
\newblock Can we edit multimodal large language models?
\newblock \emph{arXiv preprint arXiv:2310.08475}.

\bibitem[{Dai et~al.(2023)Dai, Li, Li, Tiong, Zhao, Wang, Li, Fung, and Hoi}]{instructblip}
Wenliang Dai, Junnan Li, Dongxu Li, Anthony Tiong, Junqi Zhao, Weisheng Wang, Boyang Li, Pascale~N Fung, and Steven Hoi. 2023.
\newblock Instructblip: Towards general-purpose vision-language models with instruction tuning.
\newblock \emph{Advances in neural information processing systems}, 36:49250--49267.

\bibitem[{De~Cao et~al.(2021)De~Cao, Aziz, and Titov}]{ke}
Nicola De~Cao, Wilker Aziz, and Ivan Titov. 2021.
\newblock Editing factual knowledge in language models.
\newblock \emph{arXiv preprint arXiv:2104.08164}.

\bibitem[{Du et~al.(2025)Du, Jiang, Gao, Shi, Zheng, Qi, and Li}]{mmke}
Yuntao Du, Kailin Jiang, Zhi Gao, Chenrui Shi, Zilong Zheng, Siyuan Qi, and Qing Li. 2025.
\newblock Mmke-bench: A multimodal editing benchmark for diverse visual knowledge.
\newblock \emph{arXiv preprint arXiv:2502.19870}.

\bibitem[{Fang et~al.(2024)Fang, Jiang, Wang, Ma, Jie, Wang, He, and Chua}]{alphaedit}
Junfeng Fang, Houcheng Jiang, Kun Wang, Yunshan Ma, Shi Jie, Xiang Wang, Xiangnan He, and Tat-Seng Chua. 2024.
\newblock Alphaedit: Null-space constrained knowledge editing for language models.
\newblock \emph{arXiv preprint arXiv:2410.02355}.

\bibitem[{Geva et~al.(2021)Geva, Schuster, Berant, and Levy}]{kv}
Mor Geva, Roei Schuster, Jonathan Berant, and Omer Levy. 2021.
\newblock Transformer feed-forward layers are key-value memories.
\newblock In \emph{Proceedings of the 2021 Conference on Empirical Methods in Natural Language Processing}, pages 5484--5495.

\bibitem[{Google(2025)}]{gemini}
Google. 2025.
\newblock Gemini 2.5 pro.
\newblock \emph{\url{https://deepmind.google/technologies/gemini/pro/}}.

\bibitem[{Gupta et~al.(2024)Gupta, Sajnani, and Anumanchipalli}]{emmet}
Akshat Gupta, Dev Sajnani, and Gopala Anumanchipalli. 2024.
\newblock A unified framework for model editing.
\newblock \emph{arXiv preprint arXiv:2403.14236}.

\bibitem[{Han et~al.(2025)Han, Wang, Fang, Zhao, Ma, and Chen}]{reasoning}
Tingxu Han, Zhenting Wang, Chunrong Fang, Shiyu Zhao, Shiqing Ma, and Zhenyu Chen. 2025.
\newblock Token-budget-aware llm reasoning.
\newblock In \emph{Findings of the Association for Computational Linguistics: ACL 2025}, pages 24842--24855.

\bibitem[{Hartvigsen et~al.(2023)Hartvigsen, Sankaranarayanan, Palangi, Kim, and Ghassemi}]{grace}
Tom Hartvigsen, Swami Sankaranarayanan, Hamid Palangi, Yoon Kim, and Marzyeh Ghassemi. 2023.
\newblock Aging with grace: Lifelong model editing with discrete key-value adaptors.
\newblock \emph{Advances in Neural Information Processing Systems}, 36:47934--47959.

\bibitem[{Huang et~al.(2024)Huang, Chen, Xu, Payani, and Shu}]{Hallucinations}
Baixiang Huang, Canyu Chen, Xiongxiao Xu, Ali Payani, and Kai Shu. 2024.
\newblock Can knowledge editing really correct hallucinations?
\newblock \emph{arXiv preprint arXiv:2410.16251}.

\bibitem[{Li et~al.(2025{\natexlab{a}})Li, Li, Song, Tang, and Zhou}]{li2025imrrf}
Dayang Li, Fanxiao Li, BingBing Song, Li~Tang, and Wei Zhou. 2025{\natexlab{a}}.
\newblock Imrrf: Integrating multi-source retrieval and redundancy filtering for llm-based fake news detection.
\newblock In \emph{Proceedings of the 2025 Conference of the Nations of the Americas Chapter of the Association for Computational Linguistics: Human Language Technologies (Volume 1: Long Papers)}, pages 9127--9142.

\bibitem[{Li et~al.(2025{\natexlab{b}})Li, Wu, Fu, Dong, Song, and Zhou}]{drift}
Fanxiao Li, Jiaying Wu, Tingchao Fu, Yunyun Dong, Bingbing Song, and Wei Zhou. 2025{\natexlab{b}}.
\newblock Drifting away from truth: Genai-driven news diversity challenges lvlm-based misinformation detection.
\newblock \emph{arXiv preprint arXiv:2508.12711}.

\bibitem[{Li et~al.(2025{\natexlab{c}})Li, Wu, He, and Zhou}]{cmie}
Fanxiao Li, Jiaying Wu, Canyuan He, and Wei Zhou. 2025{\natexlab{c}}.
\newblock Cmie: Combining mllm insights with external evidence for explainable out-of-context misinformation detection.
\newblock \emph{arXiv preprint arXiv:2505.23449}.

\bibitem[{Li et~al.(2026)Li, Liu, Liu, Wang, Yin, and Xu}]{Haoran}
Haoran Li, Renyang Liu, Hongjia Liu, Chen Wang, Long Yin, and Jian Xu. 2026.
\newblock Pwavep: Purifying imperceptible adversarial perturbations in 3d point clouds via spectral graph wavelets.
\newblock In \emph{Proceedings of the ACM Web Conference 2026}, WWW '26.

\bibitem[{Li et~al.(2023)Li, Li, Savarese, and Hoi}]{blip2}
Junnan Li, Dongxu Li, Silvio Savarese, and Steven Hoi. 2023.
\newblock Blip-2: Bootstrapping language-image pre-training with frozen image encoders and large language models.
\newblock In \emph{International conference on machine learning}, pages 19730--19742. PMLR.

\bibitem[{Li et~al.(2024)Li, Li, Song, Yang, Ma, and Yu}]{li2024pmet}
Xiaopeng Li, Shasha Li, Shezheng Song, Jing Yang, Jun Ma, and Jie Yu. 2024.
\newblock Pmet: Precise model editing in a transformer.
\newblock In \emph{Proceedings of the AAAI Conference on Artificial Intelligence}, volume~38, pages 18564--18572.

\bibitem[{Liu et~al.(2023)Liu, Li, Wu, and Lee}]{llava}
Haotian Liu, Chunyuan Li, Qingyang Wu, and Yong~Jae Lee. 2023.
\newblock Visual instruction tuning.
\newblock \emph{Advances in neural information processing systems}, 36:34892--34916.

\bibitem[{Liu et~al.(2025{\natexlab{a}})Liu, Feng, Zhang, Zhou, Cheng, and Ng}]{ccs}
Renyang Liu, Wenjie Feng, Tianwei Zhang, Wei Zhou, Xueqi Cheng, and See-Kiong Ng. 2025{\natexlab{a}}.
\newblock Rethinking machine unlearning in image generation models.
\newblock In \emph{Proceedings of the 2025 ACM SIGSAC Conference on Computer and Communications Security}, pages 993--1007.

\bibitem[{Liu et~al.(2025{\natexlab{b}})Liu, Lyu, Zhou, and Ng}]{liu2025}
Renyang Liu, Ziyu Lyu, Wei Zhou, and See-Kiong Ng. 2025{\natexlab{b}}.
\newblock D-judge: How far are we? assessing the discrepancies between ai-synthesized and natural images through multimodal guidance.
\newblock In \emph{Proceedings of the 33rd ACM International Conference on Multimedia}, pages 10797--10806.

\bibitem[{Meng et~al.(2022{\natexlab{a}})Meng, Bau, Andonian, and Belinkov}]{rome}
Kevin Meng, David Bau, Alex Andonian, and Yonatan Belinkov. 2022{\natexlab{a}}.
\newblock Locating and editing factual associations in gpt.
\newblock \emph{Advances in neural information processing systems}, 35:17359--17372.

\bibitem[{Meng et~al.(2022{\natexlab{b}})Meng, Sharma, Andonian, Belinkov, and Bau}]{memit}
Kevin Meng, Arnab~Sen Sharma, Alex Andonian, Yonatan Belinkov, and David Bau. 2022{\natexlab{b}}.
\newblock Mass-editing memory in a transformer.
\newblock \emph{arXiv preprint arXiv:2210.07229}.

\bibitem[{Mitchell et~al.(2021)Mitchell, Lin, Bosselut, Finn, and Manning}]{mend}
Eric Mitchell, Charles Lin, Antoine Bosselut, Chelsea Finn, and Christopher~D Manning. 2021.
\newblock Fast model editing at scale.
\newblock \emph{arXiv preprint arXiv:2110.11309}.

\bibitem[{Mitchell et~al.(2022)Mitchell, Lin, Bosselut, Manning, and Finn}]{serac}
Eric Mitchell, Charles Lin, Antoine Bosselut, Christopher~D Manning, and Chelsea Finn. 2022.
\newblock Memory-based model editing at scale.
\newblock In \emph{International Conference on Machine Learning}, pages 15817--15831. PMLR.

\bibitem[{Pan et~al.(2024{\natexlab{a}})Pan, Cao, Wang, Yang, and Wang}]{finding}
Haowen Pan, Yixin Cao, Xiaozhi Wang, Xun Yang, and Meng Wang. 2024{\natexlab{a}}.
\newblock Finding and editing multi-modal neurons in pre-trained transformers.
\newblock In \emph{Findings of the Association for Computational Linguistics: ACL 2024}, pages 1012--1037.

\bibitem[{Pan et~al.(2025)Pan, Wang, Cao, Shi, Yang, Li, and Wang}]{fine}
Haowen Pan, Xiaozhi Wang, Yixin Cao, Zenglin Shi, Xun Yang, Juanzi Li, and Meng Wang. 2025.
\newblock Precise localization of memories: A fine-grained neuron-level knowledge editing technique for llms.
\newblock \emph{arXiv preprint arXiv:2503.01090}.

\bibitem[{Pan et~al.(2024{\natexlab{b}})Pan, Fan, Li, Yu, Fei, Tang, Hong, Zhang, and Sun}]{unike}
Kaihang Pan, Zhaoyu Fan, Juncheng Li, Qifan Yu, Hao Fei, Siliang Tang, Richang Hong, Hanwang Zhang, and Qianru Sun. 2024{\natexlab{b}}.
\newblock Towards unified multimodal editing with enhanced knowledge collaboration.
\newblock \emph{Advances in Neural Information Processing Systems}, 37:110290--110314.

\bibitem[{Peng et~al.(2023)Peng, Li, He, Galley, and Gao}]{vicuna}
Baolin Peng, Chunyuan Li, Pengcheng He, Michel Galley, and Jianfeng Gao. 2023.
\newblock Instruction tuning with gpt-4.
\newblock \emph{arXiv preprint arXiv:2304.03277}.

\bibitem[{Radford et~al.(2021)Radford, Kim, Hallacy, Ramesh, Goh, Agarwal, Sastry, Askell, Mishkin, Clark et~al.}]{clip}
Alec Radford, Jong~Wook Kim, Chris Hallacy, Aditya Ramesh, Gabriel Goh, Sandhini Agarwal, Girish Sastry, Amanda Askell, Pamela Mishkin, Jack Clark, and 1 others. 2021.
\newblock Learning transferable visual models from natural language supervision.
\newblock In \emph{International conference on machine learning}, pages 8748--8763. PmLR.

\bibitem[{Touvron et~al.(2023)Touvron, Martin, Stone, Albert, Almahairi, Babaei, Bashlykov, Batra, Bhargava, Bhosale et~al.}]{llama}
Hugo Touvron, Louis Martin, Kevin Stone, Peter Albert, Amjad Almahairi, Yasmine Babaei, Nikolay Bashlykov, Soumya Batra, Prajjwal Bhargava, Shruti Bhosale, and 1 others. 2023.
\newblock Llama 2: Open foundation and fine-tuned chat models.
\newblock \emph{arXiv preprint arXiv:2307.09288}.

\bibitem[{Wang et~al.(2024)Wang, Wang, Zhou, Wang, Li, Hua, and Tang}]{understand}
Yabing Wang, Le~Wang, Qiang Zhou, Zhibin Wang, Hao Li, Gang Hua, and Wei Tang. 2024.
\newblock Multimodal llm enhanced cross-lingual cross-modal retrieval.
\newblock In \emph{Proceedings of the 32nd ACM International Conference on Multimedia}, pages 8296--8305.

\bibitem[{Wu et~al.(2025)Wu, Li, Fu, Kan, and Hooi}]{see}
Jiaying Wu, Fanxiao Li, Zihang Fu, Min-Yen Kan, and Bryan Hooi. 2025.
\newblock Seeing through deception: Uncovering misleading creator intent in multimodal news with vision-language models.
\newblock \emph{arXiv preprint arXiv:2505.15489}.

\bibitem[{Yang et~al.(2025)Yang, Li, Yang, Zhang, Hui, Zheng, Yu, Gao, Huang, Lv et~al.}]{qwen3}
An~Yang, Anfeng Li, Baosong Yang, Beichen Zhang, Binyuan Hui, Bo~Zheng, Bowen Yu, Chang Gao, Chengen Huang, Chenxu Lv, and 1 others. 2025.
\newblock Qwen3 technical report.
\newblock \emph{arXiv preprint arXiv:2505.09388}.

\bibitem[{Zheng et~al.(2023)Zheng, Li, Dong, Fan, Wu, Xu, and Chang}]{ike}
Ce~Zheng, Lei Li, Qingxiu Dong, Yuxuan Fan, Zhiyong Wu, Jingjing Xu, and Baobao Chang. 2023.
\newblock Can we edit factual knowledge by in-context learning?
\newblock \emph{arXiv preprint arXiv:2305.12740}.

\bibitem[{Zhu et~al.(2023)Zhu, Chen, Shen, Li, and Elhoseiny}]{minigpt}
Deyao Zhu, Jun Chen, Xiaoqian Shen, Xiang Li, and Mohamed Elhoseiny. 2023.
\newblock Minigpt-4: Enhancing vision-language understanding with advanced large language models.
\newblock \emph{arXiv preprint arXiv:2304.10592}.

\end{thebibliography}

\appendix

\appendix
\clearpage

\appendix

\section{Appendix}
\label{sec:appendix}

\subsection{Neuron Analysis on LLaVA and Qwen-VL}
\label{sec:llava}
\begin{figure*}
    \centering
    \includegraphics[width=0.99\linewidth]{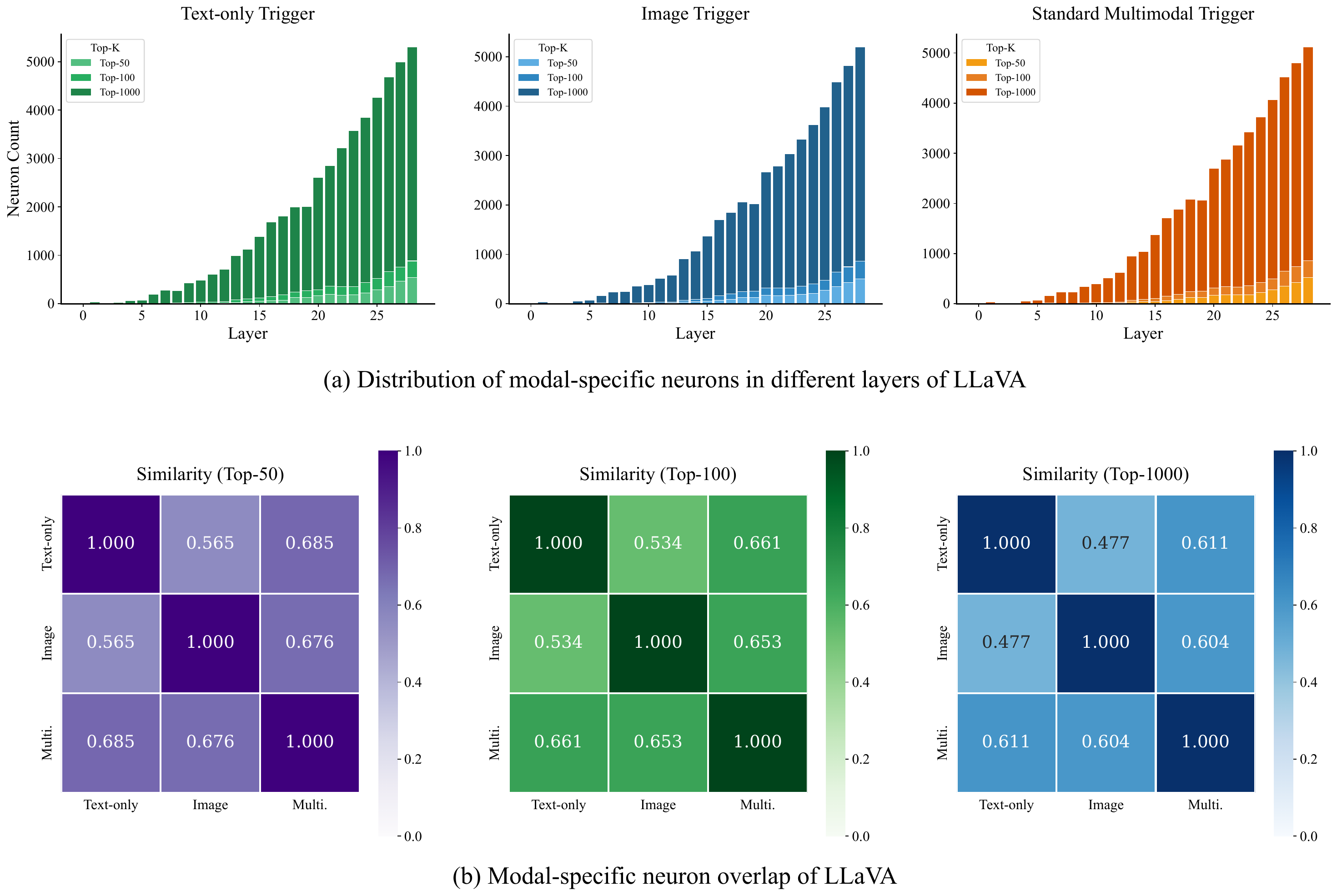}
    \caption{\textbf{Layer-wise distribution and modality overlap of critical neurons in LLaVA.} (a) Distribution of high-contribution neurons across layers under Text-only, Image, and Standard Multimodal triggers. (b) Similarity matrices measuring the overlap between modality-specific neuron sets at of top-50, 100, and 1000. Consistent with our findings on InstructBLIP, LLaVA exhibits minimal overlap between textual and visual pathways.}
    \label{fig:neuron_la}
\end{figure*}

\begin{figure*}
    \centering
    \includegraphics[width=0.99\linewidth]{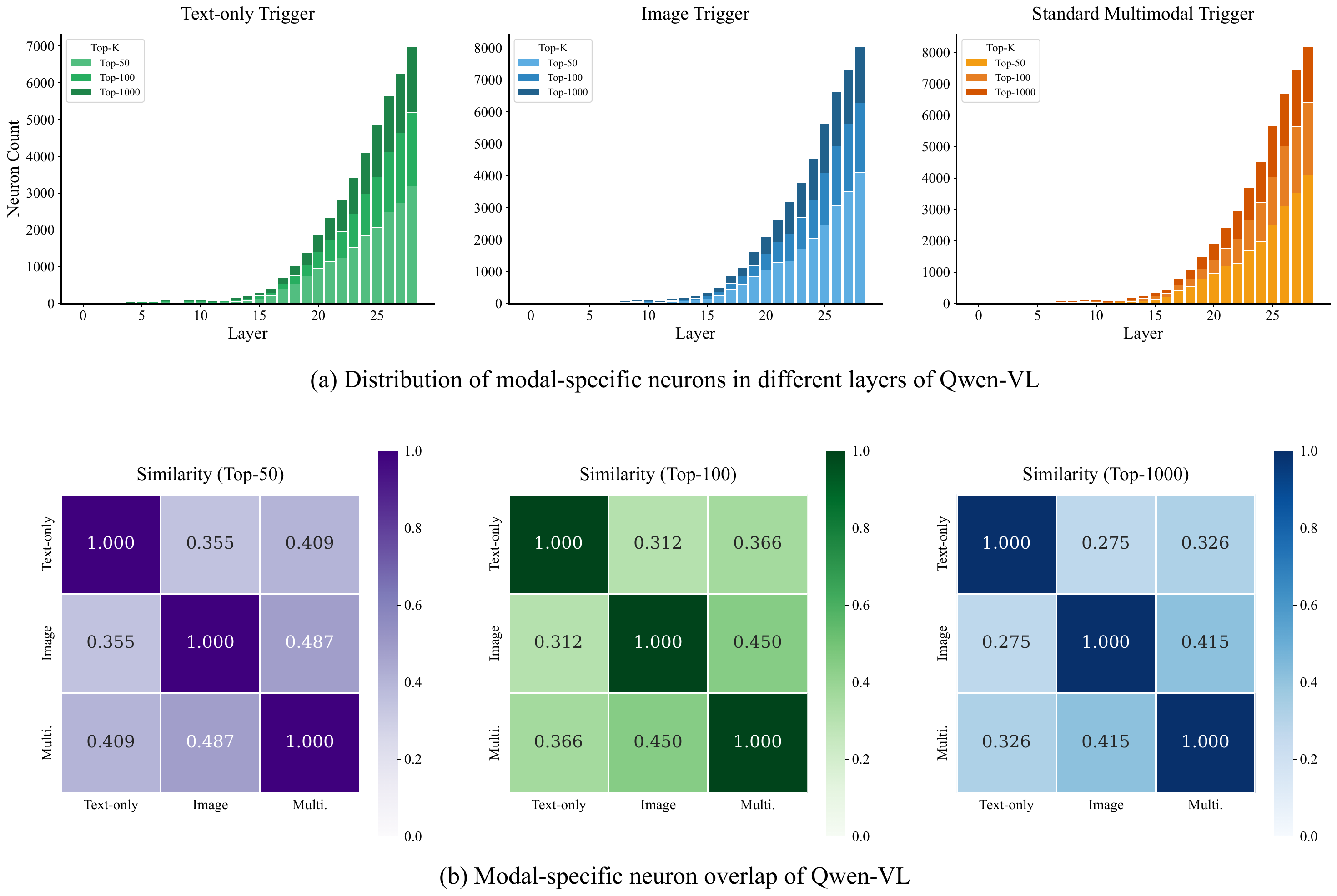}
    \caption{\textbf{Layer-wise distribution and modality overlap of critical neurons in Qwen-VL.} (a) Distribution of high-contribution neurons across layers under Text-only, Image, and Standard Multimodal triggers. (b) Similarity matrices measuring the overlap between modality-specific neuron sets at of top-50, 100, and 1000. Consistent with our findings on InstructBLIP, Qwen-VL exhibits minimal overlap between textual and visual pathways.}
    \label{fig:neuron_qwen}
\end{figure*}

To further investigate the architectural robustness of DECODE, we extend our neuron activation analysis to LLaVA-v1.5-7B and Qwen-VL-7B. While Figure~\ref{fig:neuron} illustrates the modality-specific distribution in InstructBLIP, Figure~\ref{fig:neuron_la} and Figure~\ref{fig:neuron_qwen} provide a comparative visualization of critical neurons in LLaVA and Qwen-VL, respectively.

Specifically, we identify the top-$k$ ($k \in \{50, 100, 1000\}$) high-contribution neurons under the three evaluation settings described in Section~\ref{sec:decoupling}: Text-only trigger, Image trigger, and Standard multimodal trigger. As shown in Figure~\ref{fig:neuron_la}, LLaVA exhibits a distinct pattern compared to InstructBLIP:

Unlike InstructBLIP, where visual-aware neurons ($U_v$) are distributed across shallower layers due to the Q-Former interface, LLaVA's modality-specific neurons are predominantly concentrated in the deeper layers of the LLM backbone. This alignment stems from LLaVA's late-fusion architecture, where visual features are projected directly into the language space, delaying cross-modal integration until the final reasoning stages.

\textbf{Modality Disentanglement}: Similarly, Qwen-VL also exhibits a relatively low overlap between $U_t$ and $U_v$, with the overlap remaining around $0.3$. Consistent with our core finding, the overlap between $U_t$ and $U_v$ remains remarkably low in LLaVA, confirming that \textit{editing decoupling failure} is not specific to one architecture but is a fundamental characteristic of how MLLMs store entity-related knowledge.

\subsection{Dataset Details}
\label{sec:data_details}
To systematically investigate the \textit{editing decoupling failure} in MLLMs, we construct a decoupled multimodal knowledge editing dataset based on the MMKE benchmark \cite{mmke}. This section details the data generation pipeline, the schema of our dataset, and the rigorous quality control measures employed.

\subsubsection{Data Construction and Decoupling Strategy}
Traditional multimodal editing benchmarks often rely on joint cues where text and images provide redundant information. To isolate modality-specific pathways, we leverage Gemini-2.5-pro \cite{gemini} to re-synthesize and decouple the original samples into three distinct evaluation settings:

\begin{itemize}
    \item \textbf{Text-only Trigger}: We extract the entity name (Subject) and generate purely linguistic queries. All visual references are removed to ensure the model can only retrieve the updated fact through its language backbone.
    \item \textbf{Image Trigger}: As shown in the prompt template (\textbf{Figure \ref{prompt:gemini}}), we explicitly instruct the LLM to remove any mention of the entity name in the query. Instead, the query refers to the target via indicative expressions (e.g., ``the person in the photo''). This forces the MLLM to rely solely on visual grounding to identify the entity.
    \item \textbf{Standard Multimodal Trigger}: A combination of both modalities with explicit entity names and corresponding images, serving as a baseline for traditional editing performance.
\end{itemize}

\subsubsection{Human Evaluation and Quality Control}
To ensure the reliability of the automated generation process, we conducted a rigorous human study. Three experts with backgrounds in NLP were tasked with evaluating a random subset of the generated data.    

The evaluation focused on the alignment between the images and entity-related text—including both explicit entity text (Text-only trigger and Standard multimodal trigger in Section~\ref{sec:data}) and implicit entity text (Image trigger in Section~\ref{sec:data}). Furthermore, we validate the semantic correctness of the rephrased queries to ensure the overall consistency and accuracy of the dataset.

As reported in Table \ref{tab:annotation_quality}, our dataset demonstrates high quality. The average accuracy across all triggers reached $96.77\%$ with a mean $\kappa$ of $93.32$.

\begin{table*}[t]
\renewcommand\arraystretch{1.1}
\setlength{\tabcolsep}{15pt}
\small
\begin{center}
\scalebox{0.94}{
\begin{tabular}{c|ccc|c}
\hline \hline
\multirow{2}{*}{\textbf{Metrics}} & \multicolumn{3}{c|}{\textbf{Annotation Consistency}} & \multirow{2}{*}{\textbf{Avg.}} \\
\cline{2-4}
 & \textbf{Text-only Trigger} & \textbf{Image Trigger} & \textbf{Rephrase Text} & \\
\hline
Accuracy (\%) & 97.00 & 95.30 & 98.00 & 96.77 \\
Fleiss'~$\kappa$ & 93.63 & 90.49 & 95.84 & 93.32 \\
\hline \hline
\end{tabular}}
\caption{\textbf{Data annotation quality and consistency.} We report the accuracy and Fleiss'~$\kappa$ scores across three different input triggers to ensure the reliability of the constructed multimodal editing dataset.}
\label{tab:annotation_quality}
\end{center}
\end{table*}

\begin{figure*}[!h]
    \centering
    \includegraphics[width=0.99\linewidth]{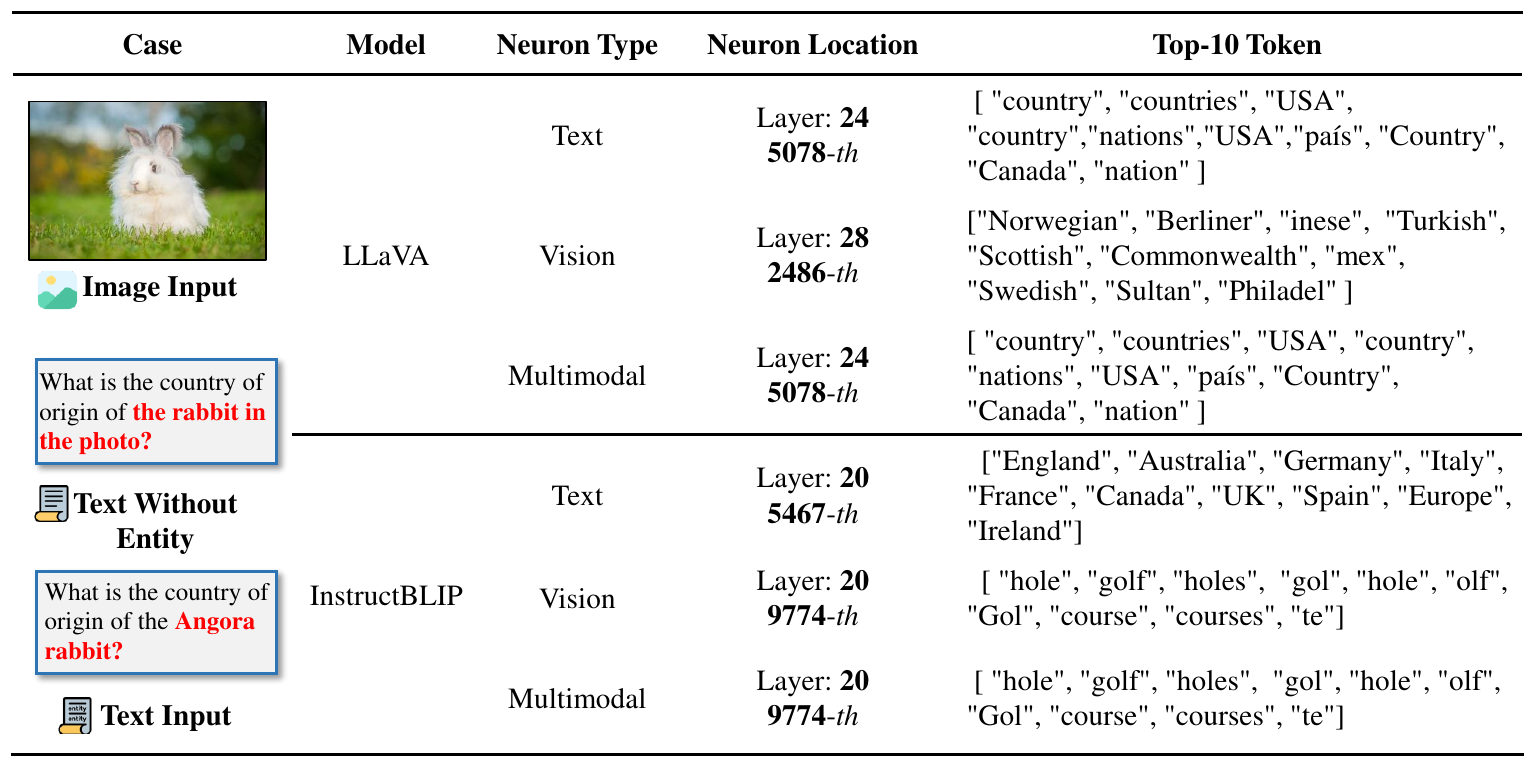}
    \caption{\textbf{Visualization of Modality-Specific Neuron Activations.} We compare the contribution scores of decoupled neuron sets across textual, visual, and multimodal queries. The results verify the functional specialization of identified neurons and their distinct distribution patterns in InstructBLIP and LLaVA.}
    \label{fig:vis}
\end{figure*}

\subsection{Editing Details}
For the implementation of DECODE, we adopt the dual-stream optimization objective defined in Section~\ref{sec:decode}. To balance the trade-off between knowledge update efficacy and model stability, the locality coefficient $\lambda$ is set to $0.6$ for both LLaVA and Qwen-VL, and $0.2$ for InstructBLIP. We utilize the Adam optimizer with a learning rate of $2 \times 10^{-3}$ to minimize the total loss. We set top-$k=40$ neurons for LLaVA and Qwen-VL, and top-$k=20$ neurons for InstructBLIP. To maintain high surgical precision and prevent over-fitting to the edit facts, we implement an early-stopping mechanism: the optimization process for each modality-specific stream is terminated once the target loss $\mathcal{L}_{tgt}$ falls below predefined thresholds ($0.3$ for the text stream and $0.7$ for the image stream in LLaVA and Qwen-VL; $0.1$ for the text stream and $0.5$ for the image stream in InstructBLIP), or when the maximum number of iterations (set to $200$) is reached.

\subsection{Case Study}
\label{sec:success_case}
\begin{figure*}
    \centering
    \includegraphics[width=0.99\linewidth]{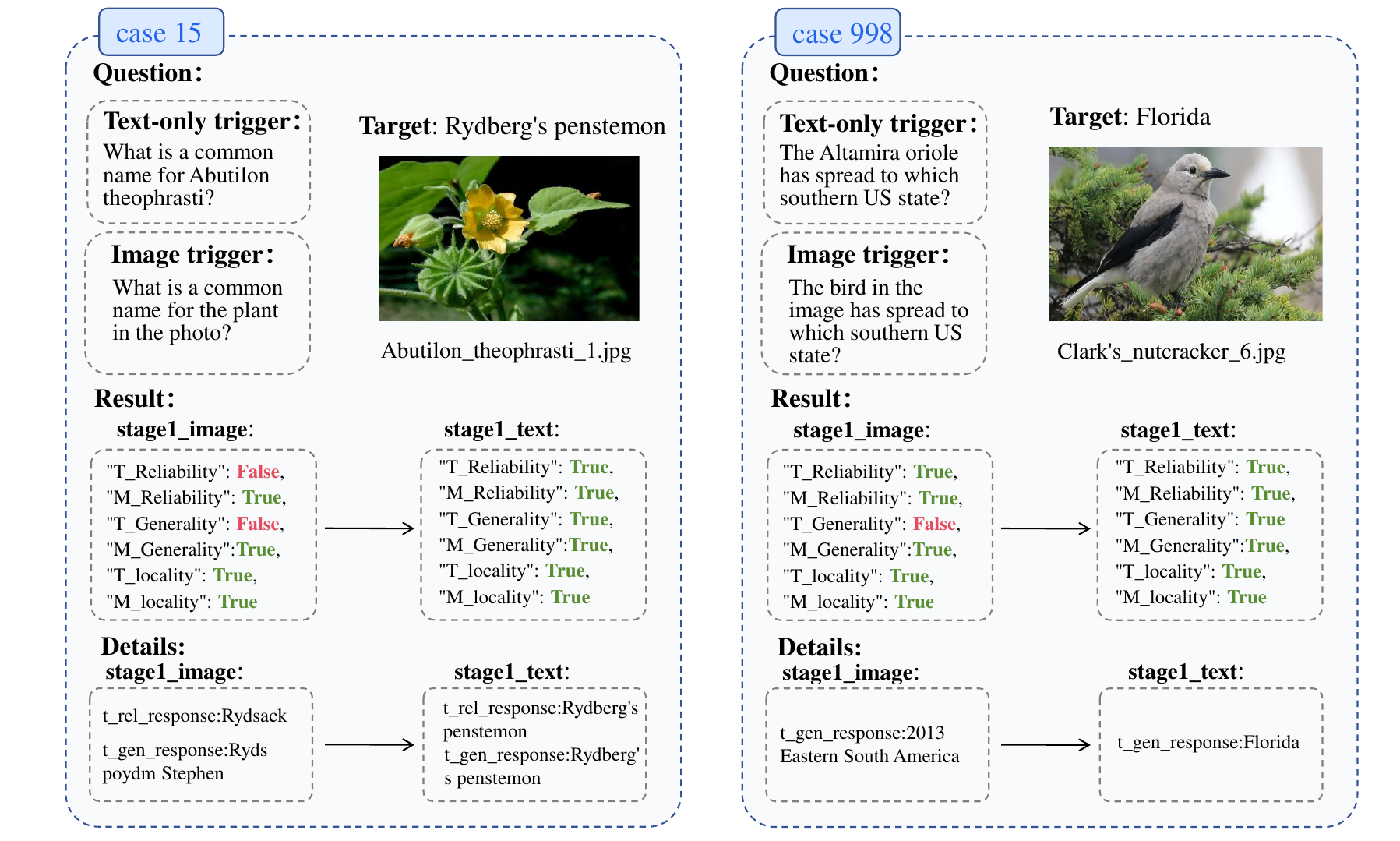}
    \caption{Success Case}
    \label{fig:success_case}
\end{figure*}

To demonstrate the effectiveness of DECODE in mitigating editing decoupling failure, we present a detailed success case in Figure~\ref{fig:success_case}. 
In this example (Case 15), the goal is to update the factual knowledge regarding the common name of the plant \textit{Abutilon theophrasti}. Specifically, the model is edited to identify the plant as \textit{"Rydberg's penstemon"} instead of its original name.

DECODE successfully synchronizes the update across pathways. When the user queries with the text-only prompt \textit{"What is a common name for Abutilon theophrasti?"}, the model correctly retrieves the edited fact \textit{"Rydberg's penstemon"}. Similarly, when queried with the image of the plant (\textit{"What is a common name for the plant in the photo?"}), the model also consistently outputs \textit{"Rydberg's penstemon"}. 
This case validates that by explicitly targeting and offsetting both visual-specific and text-specific neuron groups ($U_v$ and $U_t$), DECODE ensures that the new knowledge is robustly accessible regardless of the input modality, effectively resolving the decoupling issue.

\subsection{Failure Analysis}
\label{sec:failure_analysis}
\begin{figure*}
    \centering
    \includegraphics[width=0.99\linewidth]{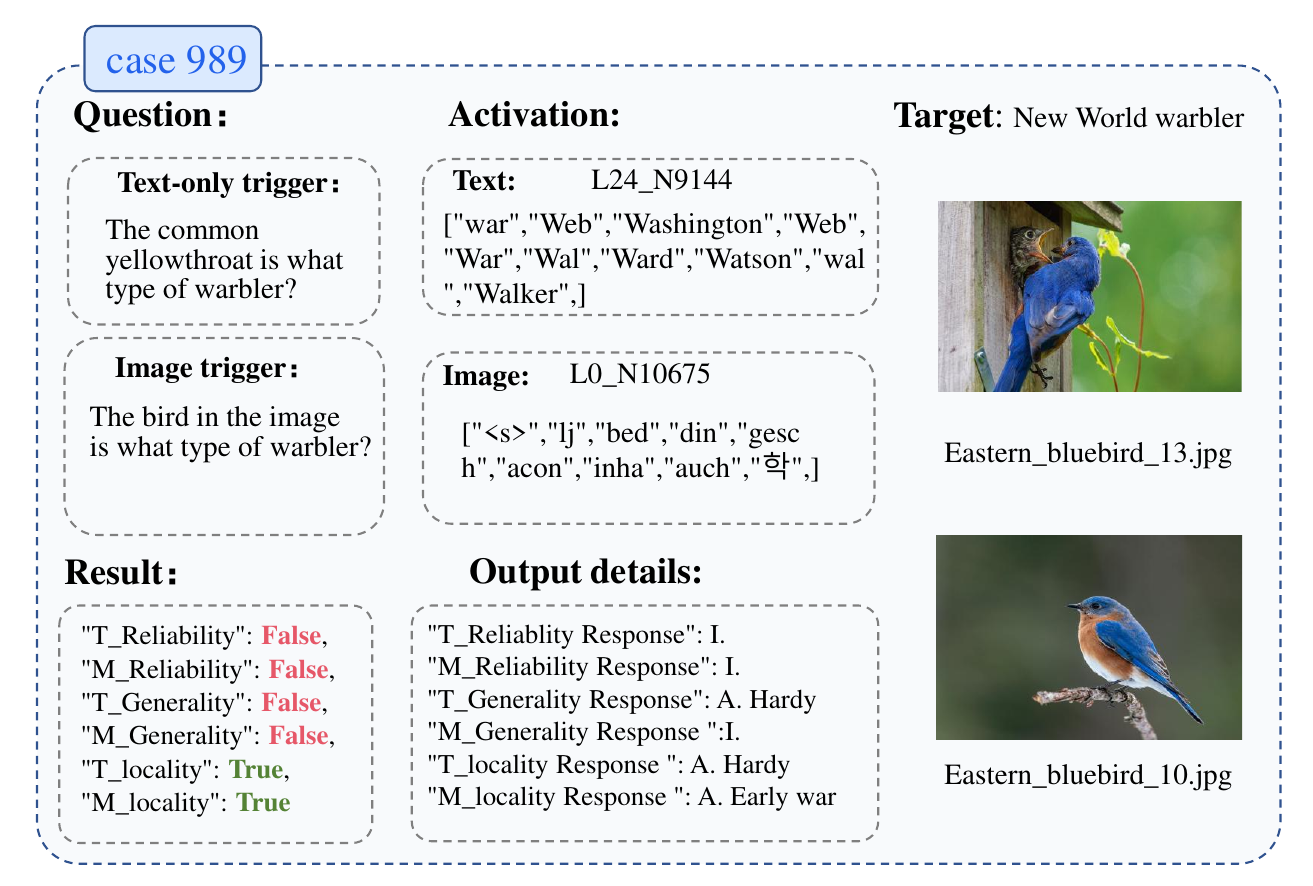}
    \caption{Failure Case}
    \label{fig:failure_case}
\end{figure*}

Despite the superior performance of DECODE, we observe specific failure patterns, as illustrated in Figure~\ref{fig:failure_case}. These failures provide insight into the boundaries of neuron-level editing in multimodal contexts. We categorize them into two main types:

\paragraph{(1) Lexical editing without semantic update.}
In some cases, the highest-contribution neurons selected for the text stream correlate more with \emph{surface-form regularities} than with the intended factual association.
Figure~\ref{fig:failure_case} (Case~989) exemplifies this phenomenon: the edit target is the taxonomy fact that \textit{Common yellowthroat} is a \textit{New World warbler}, yet the identified text-critical neuron is primarily associated with tokens sharing the \texttt{War-} prefix (e.g., ``war'', ``War'', ``Ward'', ``Walker'').
As a result, optimizing offsets on such neurons tends to modify the model's \emph{lexical handling} of \texttt{War-}-initial subwords rather than updating the underlying semantic relation ``\textit{Common yellowthroat} $\rightarrow$ \textit{New World warbler}''.
This lexical-level intervention can appear brittle: when the query is rephrased or the generation pathway does not strongly rely on the same prefix-sensitive circuit, the edited knowledge is not reliably retrieved, leading to degraded reliability and generality while leaving locality largely intact.

\paragraph{(2) Weak grounding yields mis-localization in the visual stream.}
A second failure mode arises when the image-triggered query provides insufficient or noisy grounding signals, causing the visual stream to activate neurons that are not truly entity-specific.
In Figure~\ref{fig:failure_case}, the image-trigger setting requires the model to infer the entity solely from ``the bird in the image'', which depends on fine-grained visual recognition and stable cross-modal alignment.
When grounding is weak (e.g., visually confusable species, background bias, or ambiguous cues), the top-ranked ``visual-critical'' neurons may shift toward generic, low-level, or non-semantic activations (often in shallower layers), producing gradients that do not consistently point toward the intended factual update.
Under our locality-regularized objective, the optimization then favors minimal, non-disruptive parameter changes, which preserves locality but fails to enforce the new fact across triggers.
This explains why, in such cases, both unimodal and multimodal evaluations can remain incorrect even after editing, revealing a practical limitation: DECODE can only synchronize knowledge across pathways when both streams successfully localize neurons that genuinely encode the target entity-relation.

\subsection{Impact of Top-$k$ Neuron Selecting}
\label{sec:ablation_k}
The choice of the top-$k$ (the number of edited neurons) is critical for balancing editing efficacy and model stability. As shown in Table~\ref{tab:topk}, there is a clear positive correlation between $k$ and the editing success rates (Reliability and Generality). Increasing $k$ allows the method to modify a larger portion of the distributed knowledge representation, ensuring the update is retrieved under diverse prompts. However, this comes at the cost of Locality. As $k$ increases, the scope of the parameter modification widens, increasing the risk of interfering with unrelated knowledge. For instance, in InstructBLIP, setting $k$ too high results in a noticeable drop in locality scores.

\begin{table*}[t]
\renewcommand\arraystretch{1.1}
\setlength{\tabcolsep}{10pt}
\small
\begin{center}
\begin{tabular}{c|c|c|cc|cc|cc|c}
\hline \hline
\multirow{2}{*}{\textbf{Model}} &
\multirow{2}{*}{\textbf{Setting}} &
\multirow{2}{*}{\textbf{Stage}} &
\multicolumn{2}{c|}{\textbf{Reliability}} &
\multicolumn{2}{c|}{\textbf{Generality}} &
\multicolumn{2}{c|}{\textbf{Locality}} &
\multirow{2}{*}{\textbf{Avg.}} \\
\cline{4-9}
& & & \textbf{T} & \textbf{M} & \textbf{T} & \textbf{M} & \textbf{T} & \textbf{M} & \\
\hline

\multirow[c]{4}{*}{InstructBLIP}
& \multirow{2}{*}{Text First}  & Stage 1 & 89.20 & 61.30 & 65.80 & 60.80 & 57.54 & 59.06 & 65.62 \\
&                               & Stage 2 & 87.22 & 83.89 & 76.73 & 83.40 & 45.78 & 46.68 & 70.62 \\
\cline{2-3}
& \multirow{2}{*}{Image First} & Stage 1 & 86.85 & 84.14 & 76.79 & 83.64 & 45.75 & 46.86 & 70.67 \\
&                               & Stage 2 & 97.79 & 92.16 & 84.81 & 91.42 & 45.45 & 46.11 & 76.29 \\
\hline

\multirow[c]{4}{*}{LLaVA}
& \multirow{2}{*}{Text First}  & Stage 1 & 74.63 & 35.43 & 64.57 & 35.74 & 72.50 & 72.41 & 59.21 \\
&                               & Stage 2 & 93.27 & 91.36 & 87.47 & 90.43 & 54.31 & 55.77 & 78.77 \\
\cline{2-3}
& \multirow{2}{*}{Image First} & Stage 1 & 65.29 & 88.82 & 59.48 & 85.61 & 46.73 & 51.37 & 66.22 \\
&                               & Stage 2 & 91.54 & 92.71 & 83.45 & 90.92 & 47.26 & 51.56 & 76.26 \\


\hline \hline
\end{tabular}
\caption{\textbf{Ablation study on editing sequences measured by Semantic Match (SM).} We report text-only (T) and multimodal (M) scores across two editing stages, demonstrating the architectural sensitivity of InstructBLIP and LLaVA to the order of modality updates.}
\label{tab:sequence_SM}
\end{center}
\end{table*}

\begin{table*}[t]
\renewcommand\arraystretch{1.1}
\setlength{\tabcolsep}{5pt}
\small
\begin{center}
\scalebox{0.94}{
\begin{tabular}{c|c|cccc|cccc|cc|cc}
\hline \hline
\multirow{3}{*}{\textbf{Model}} &
\multirow{3}{*}{\textbf{Method}} &
\multicolumn{4}{c|}{\textbf{Reliability}} &
\multicolumn{4}{c|}{\textbf{Generality}} &
\multicolumn{2}{c|}{\textbf{Locality}} &
\multicolumn{2}{c}{\textbf{Avg.}} \\
\cline{3-14}
& & \multicolumn{2}{c|}{\textbf{T}} & \multicolumn{2}{c|}{\textbf{M}} & \multicolumn{2}{c|}{\textbf{T}} & \multicolumn{2}{c|}{\textbf{M}} & \multirow{2}{*}{\textbf{T}} & \multirow{2}{*}{\textbf{M}} & \multirow{2}{*}{\textbf{EM}} & \multirow{2}{*}{\textbf{SM}} \\
\cline{3-10}
& & \textbf{EM} & \textbf{SM} & \textbf{EM} & \textbf{SM} & \textbf{EM} & \textbf{SM} & \textbf{EM} & \textbf{SM} & & & & \\
\hline

\multirow[c]{5}{*}{InstructBLIP}
& top-10 & 90.11 & 90.00 & 65.17 & 64.00 & 47.80 & 47.00 & 61.47 & 58.00 & 55.93 & 64.63 & 64.19 & 63.26 \\
& top-20 & 95.00 & 95.00 & 84.93 & 87.00 & 87.86 & 82.00 & 86.00 & 87.0 & 40.74 & 44.96 & 73.25 & 72.78 \\
& top-30 & 98.10 & 97.00 & 98.10 & 98.00 & 87.50 & 89.00 & 96.50 & 97.00 & 44.23 & 39.93 & 77.39 & 77.53 \\
& top-40  & 98.50 & 98.00 & 99.10 & 98.00 & 88.80 & 93.00 & 97.75 & 97.00 & 37.35 & 43.53 & 77.51 & 77.81 \\
& top-50  & 99.30 & 99.00 & 87.31 & 85.00 & 79.73 & 80.00 & 86.20 & 84.00 & 45.60 & 50.71 & 74.81 & 74.05 \\
\hline

\multirow[c]{5}{*}{LLaVA}
& top-10    & 93.50 & 92.00 & 95.83 & 94.00 & 88.25 & 87.00 & 91.33 & 89.00 & 34.63 & 37.02 & 73.43 & 72.28 \\
& top-20 & 98.44 & 98.00 & 91.75 & 91.00 & 94.00 & 94.00 & 89.61 & 88.00 & 45.36 & 49.77 & 78.16 & 77.69 \\
& top-30   & 97.53 & 96.00 & 93.19 & 91.00 & 92.03 & 94.00 & 91.19 & 88.00 & 46.52 & 52.32 & 78.80 & 77.64 \\
& top-40  & 99.20 & 99.00 & 97.67 & 97.00 & 95.00 & 95.00 & 95.42 & 94.00 & 44.55 & 49.77 & 80.27 & 79.89 \\
& top-50  & 98.76 & 98.00 & 95.94 & 94.00 & 92.76 & 92.00 & 95.08 & 94.00 & 37.18 & 42.14 & 76.98 & 76.22 \\

\hline \hline
\end{tabular}}
\caption{Impact of \textbf{top-$k$ neuron selection} on editing performance across different modalities. We investigate how varying the number of edited neurons ($k$) affects metrics under \textbf{Exact Match (EM)} and \textbf{Semantic Match (SM)}. Locality is determined by the cosine similarity of outputs before and after editing, while \textbf{Avg.} columns integrate performance across all primary dimensions.}
\label{tab:topk}
\end{center}
\end{table*}

\begin{figure*}
    \centering
    \includegraphics[width=1.0\linewidth]{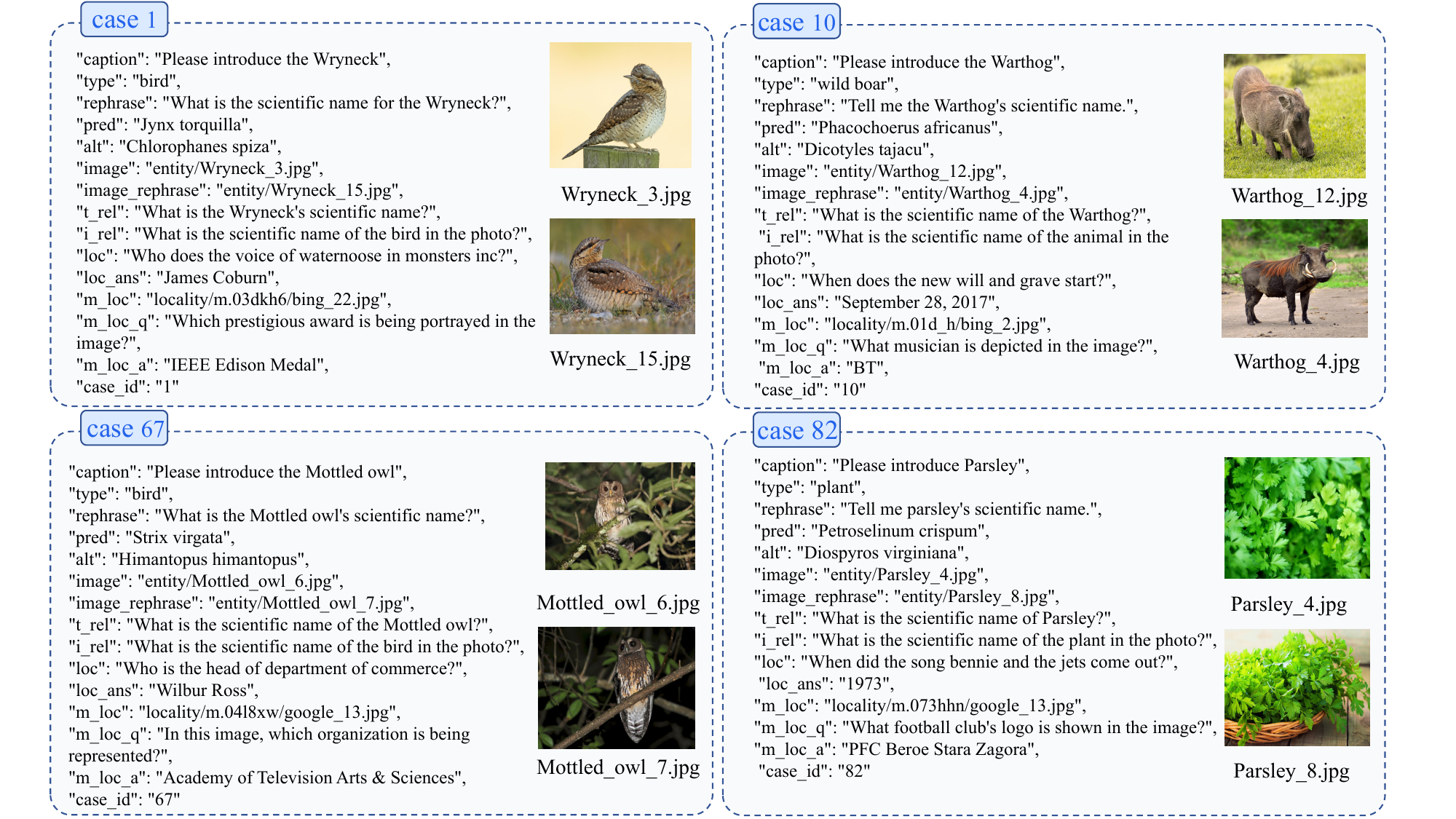}
    \caption{\textbf{Illustrative examples from our decoupled multimodal knowledge editing dataset.} We present four representative cases (\textit{case 1}, \textit{case 10}, \textit{case 67} and \textit{case 82}) showcasing the structured metadata, target knowledge (e.g., Rydberg's penstemon, Florida), and the decoupling of queries. Each sample includes a textual trigger (\textit{t\_rel}) and a vision-based trigger (\textit{i\_rel}) designed to evaluate whether the MLLM can consistently retrieve the updated fact across different modality pathways.}
    \label{fig:dataset_case}
    \label{fig:data_detail}
\end{figure*}


\begin{figure*}
\begin{prompt}{Gemini Prompt}
\small  

This is the dataset before modification \{......\},
This is the modified dataset I need.
\begin{verbatim}
{
  "caption": "Please introduce Cher",
  "type": "human",
  "rephrase": "What is Cher's original name?",
  "pred": "Cherilyn Sarkisian",
  "alt": "Steven Jay Smith",
  "image": "entity/Cher+6.jpg",
  "image_rephrase": "entity/Cher+19.jpg",
  "t_rel": "What is Cher's birth name?",
  "i_rel": "What is the original name of the person in the photo?",
  "loc": "What are the accounting standards issued by the iasb called?",
  "loc_ans": "The International Financial Reporting Standards...",
  "m_loc": "locality/m.04b2qn/google_12.jpg",
  "m_loc_q": "What film is shown in the image?",
  "m_loc_a": "Sideways"
},
\end{verbatim}
 I will continue to provide you with the original dataset later. Please help me modify it to the one I need. Please note that the purpose of me building this new dataset here is to explore whether modifying one of the image modalities and text modalities for the same concept (in the example, it is the original name of cher) will affect the other. Therefore, For image issues, i\_rel should not include should not include the concept name cher that we need to modify. 

\end{prompt}
\caption{Prompt for Dataset Construction.}
\label{prompt:gemini}
\end{figure*}

\end{document}